\documentclass[11pt]{article}
\usepackage{acl}

\usepackage{times}
\usepackage{latexsym}
\usepackage[T1]{fontenc}
\usepackage[utf8]{inputenc}
\usepackage{microtype}
\usepackage{inconsolata}
\usepackage{graphicx}
\usepackage{placeins}
\usepackage{soul}

\usepackage{hyperref}
\usepackage{url}

\usepackage{amsmath,amsfonts,bm}

\def\eqref#1{equation~\ref{#1}}

\def\1{\bm{1}}

\DeclareMathAlphabet{\mathsfit}{\encodingdefault}{\sfdefault}{m}{sl}
\SetMathAlphabet{\mathsfit}{bold}{\encodingdefault}{\sfdefault}{bx}{n}

\usepackage{amsmath,amsfonts,amssymb}
\usepackage{nicefrac}
\usepackage{booktabs}
\usepackage{multirow}
\usepackage{array}
\usepackage{longtable}
\usepackage{colortbl}
\usepackage{float}
\usepackage{xspace}
\usepackage{pifont}
\usepackage{tikz}
\usepackage{tcolorbox}
\tcbuselibrary{breakable}
\usepackage{wrapfig}
\usepackage{xcolor}

\definecolor{bluecolorone}{HTML}{AFCBF3}
\definecolor{bluecolortwo}{HTML}{6195C5}
\definecolor{yellowcolor}{HTML}{DCD494}
\definecolor{orangecolor}{HTML}{E49E7A}
\definecolor{redcolor}{HTML}{E39183}
\definecolor{casestudycolor}{HTML}{1E90FF}
\definecolor{almond}{rgb}{0.94, 0.87, 0.8}
\definecolor{lightblue}{RGB}{240,248,255}
\definecolor{azurecw}{rgb}{0.0, 0.5, 1.0}
\definecolor{blush}{rgb}{0.87, 0.36, 0.51}

\definecolor{catA}{RGB}{235, 245, 255}   
\definecolor{catB}{RGB}{255, 243, 235}   
\definecolor{catC}{RGB}{237, 252, 243}   
\definecolor{nodecolor}{RGB}{60, 80, 160}
\definecolor{cellshade}{HTML}{F0EDFF}

\newtcolorbox{theobox}{
    breakable,
  colback=casestudycolor!10,
  colframe=casestudycolor!10,
  coltitle=black,
  fonttitle=\bfseries,
  boxrule=0.5pt,
  arc=0pt,
  left=3pt, right=3pt, top=3pt, bottom=3pt,
}

\newenvironment{changemargin}[2]{\begin{list}{}{
  \setlength{\topsep}{0pt}\setlength{\leftmargin}{0pt}
  \setlength{\rightmargin}{0pt}
  \setlength{\listparindent}{\parindent}
  \setlength{\itemindent}{\parindent}
  \setlength{\parsep}{0pt plus 1pt}
  \addtolength{\leftmargin}{#1}\addtolength{\rightmargin}{#2}
  }\item}{\end{list}}

\newcommand{\lightmidrule}{\arrayrulecolor{gray!50}\midrule\arrayrulecolor{black}}

\newcommand{\bench}{\textsc{EstBook}\xspace}
\newcommand{\nqtype}{29\xspace}
\newcommand{\ndata}{10,576\xspace}

\title{From Test-taking to Cognitive Scaffolding: A Pedagogical Diagnostic Benchmark for LLMs on English Standardized Tests}

\author{
Luoxi Tang$^{1}$, Tharunya Sundar$^{1}$, Yuqiao Meng$^{1}$, Shuai Yang$^{1}$, Ankita Patra$^{1}$,\\
\textbf{Lakshmi Manohar Chippada$^{1}$, Jiqian Zhao$^{2}$, Yi Li$^{2}$, Weicheng Ma$^{3}$, Zhaohan Xi$^{1}$}\\[2mm]
$^{1}$Binghamton University, $^{2}$BlossomsAI, $^{3}$Dartmouth College\\
\texttt{\{ltang24,zxi1\}@binghamton.edu, \{jiqian,yi\}@blossoms.ai}
}

\begin{document}
\maketitle

\begin{abstract}
  As large language models (LLMs) are increasingly integrated into educational tools, current evaluations on standardized tests predominantly focus on binary outcome accuracy. Instead, an effective AI tutor must exhibit faithful reasoning, elucidate solution strategies, and diagnose specific human misconceptions. To bridge this gap, we introduce a pedagogical diagnostic framework that models English Standardized Test (EST) problem-solving as a traversal through a cognitive framework. Based on this framework, we present \bench, a multimodal benchmark encompassing \ndata questions and \nqtype task types across five major exams. Unlike traditional datasets, \bench goes beyond data aggregation by enriching questions with formalized reasoning trajectories and distractor rationales that capture specific cognitive traps. Through extensive evaluations, we empirically demonstrate the practical utility of our diagnostic framework, showing that identifying cognitive trajectories facilitates the mitigation of performance gap and improves pedagogical reasoning through guided elicitation.
\end{abstract}

\section{Introduction}



AI-driven tools are rapidly transforming the education industry, with large language models (LLMs) increasingly integrated into \textit{English Standardized Tests} (ESTs) such as TOEFL, IELTS, and GRE. Recent advances highlight the use of LLMs in automated scoring \citep{xia2024empirical,zhong2024evaluating,gupta2023testing}, test preparation \citep{feng2024exploring,ashrafimoghari2024evaluating}, and question generation \citep{tiratatri2025designing}. However, the majority of current research evaluating LLMs on standardized tests focuses exclusively on outcome metrics, i.e., whether the model ultimately performs the correct answers.

Notably, raw answering accuracy is fundamentally insufficient for deploying LLMs as intelligent educational assistants. In pedagogical practice, a tutor must not only solve the problem but also \textbf{(i)} exhibit a rigorous and faithful reasoning trajectory \citep{xu2025explainable,worden2026foundationalassist,yang2025eligibility}, \textbf{(ii)} generate explanations that clearly elucidate the solution strategy align with human practice \citep{abdulsalam2025large,meng2025benchmarking}, and \textbf{(iii)} articulate why distractive options are incorrect by diagnosing the specific human misconceptions they represent \citep{liu2025llms,mitton2026misconception,hooshyar2025problems}.
An LLM that arrives at the correct answer through flawed intermediate logic, or one that cannot explain why a specific distractive option represents a common cognitive trap, possesses high answering capability but zero pedagogical utility. To elicit LLMs from simple test-takers to effective tutors, we must evaluate their problem-solving trajectories rather than just their final outputs.

 \begin{figure}[!tp]
    \centering
    \includegraphics[width=\columnwidth]{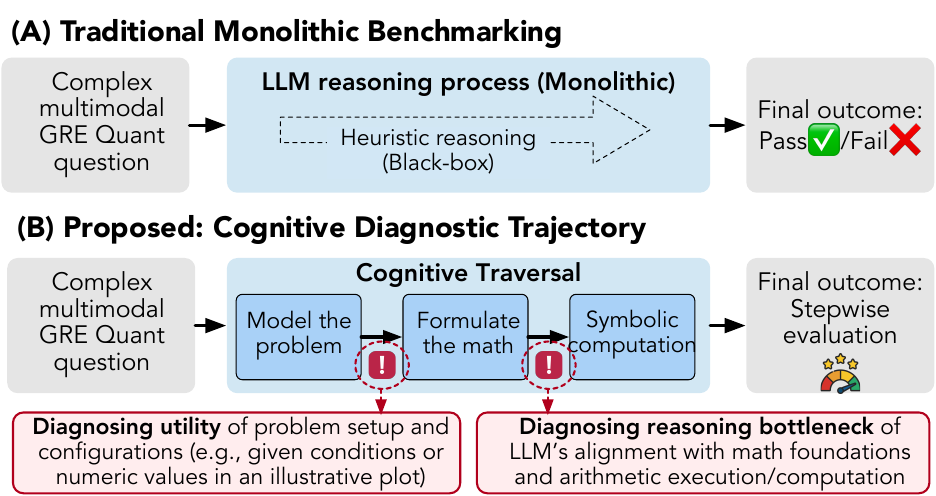}
    \caption{Comparison between monolithic LLM reasoning and our proposed pedagogical diagnostics.}
    \label{fig:intro}
\end{figure}

To bridge this gap, we introduce a formalized pedagogical diagnostic benchmark that models standardized test-taking as a \textbf{Cognitive Trajectory}. Motivated by the principle of cognitive diagnosis models \citep{leighton2007cognitive} and the common practice of evidence-centered designs using across test centers \citep{mislevy2003brief}, we highlight that answering any complex EST question is not a monolithic task, but a traversal through a structured graph of cognitive sub-skills. As illustrated in Figure \ref{fig:intro}, solving a multimodal GRE quantitative question requires a progression from modeling the problem (parsing variables from a chart), to mathematical formulation (formulating an equation), to symbolic computation (calculation). By formalizing these steps, our framework allows us to isolate exact reasoning bottlenecks to identify precisely which step in the cognitive trajectory causes an LLM (or a participant) to fail.

To empirically validate this framework, we introduce \bench, a novel pedagogical diagnostic benchmark spanning five internationally recognized ESTs (TOEFL, IELTS, SAT, GRE, and GMAT). \bench encompasses \nqtype question types and \ndata multimodal examples (incorporating text, images, tables, and mathematical symbols). Notable, our primary contribution is not merely data aggregation. Instead, we heavily manipulate and annotate the dataset for diagnostic purpose. Beyond ground-truth answers, \bench is enriched with structured breakdown steps mapping to our cognitive trajectory, as well as distractor options that detail the specific cognitive traps or misconceptions associated with incorrect options.


Finally, we conduct extensive empirical evaluations to demonstrate the diagnostic and practical value of \bench. We utilize the cognitive trajectories, evaluate the stepwise performance, and reveal vulnerabilities where LLMs are highly susceptible to cognitive steps or distractors. Furthermore, we demonstrate that studying these cognitive trajectories facilitates the design of highly effective, task-specific mitigation (e.g., evidence-anchored elicitation and symbolic verification). We  enforce strict logical reasoning exactly where the models typically hallucinate to significantly improve LLM robustness and reasoning performance across both knowledge-intensive, reasoning-heavy , or integrative pedagogical tasks.

In summary, our contributions are threefold: 

\textbf{(1) \bench: } We formalize a  cognitive diagnostic benchmark, \bench, contains pedagogical annotations and distractor rationales that shifts the evaluation of LLMs on English tests from binary accuracy to step-by-step cognitive reasoning. 

\textbf{(2) Comprehensive coverage:} \bench covers diverse modalities, containing \nqtype standardized test types including \ndata questions to support comprehensive evaluation purposes. 

\textbf{(3) Diagnostic evaluation:} We empirically evaluate the effectiveness of our approach across different LLMs. Furthermore, we demonstrate how our framework can directly guide improvements of model reasoning, which highlights the practical utility for educational practitioners. Dataset and code are available at: \url{https://anonymous.4open.science/r/Education-9595}.

\section{Related Work}

{\bf LLMs for Education.} LLMs are increasingly used in education for tasks like grading \citep{chang2024survey,holmes2022state}, question generation \citep{,mohebbi2025enabling,zhang2024llama}, and tutoring \citep{schmucker2024ruffle}. Early systems like Codex showed that LLMs can help students across language learning by offering contextual feedback and answers \citep{chiang2024large,liang2024improving,wen2024ai}. More recent research explores using LLMs as conversational tutors \citep{sabri2025performance}. However, most existing evaluations rely on limited benchmarks or informal studies, often focusing on narrow skills like math or reading \citep{guilherme2019ai,lee2023generative}. 


{\bf Benchmarking LLMs.} Many benchmarks test LLMs on general or domain-specific reasoning tasks \citep{yang2025eligibility,cao2025toward,meng2025benchmarking,meng2025uncovering}, evaluating their factual knowledge and reasoning abilities. Some, like MathVista \citep{lu2023mathvista,peng2024multimath} and ScienceQA \citep{wen2024characterizing,zhang2024simulating}, include images or structured data, but often use synthetic problems or cover narrow domains. Our benchmark, \bench, is grounded in real standardized exams and spans multiple formats (e.g., multi-choice, text completion), providing a more realistic test of LLMs as educational agents in a heterogeneous problem-solving environment.

\begin{figure*}[!tp]
    \centering
    \includegraphics[width=\textwidth]{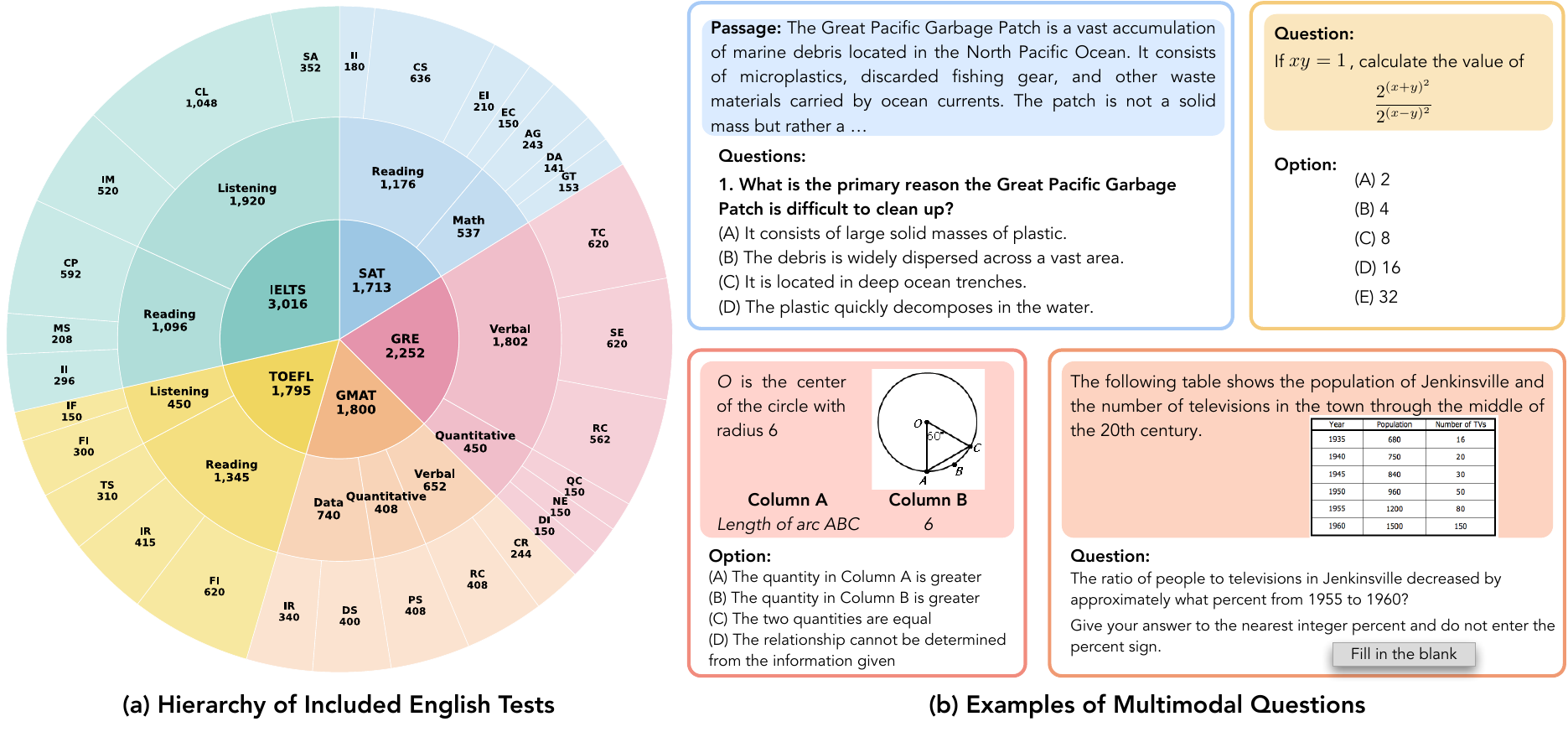}
    \caption{\textbf{Overview of \bench}: (a) the hierarchical structure of included English tests (detailed descriptions and abbreviations are provided in Table \ref{tab:benchmark}, with concrete examples in Appendix \ref{app:q-type}). (b) Examples of mutimodal questions in \bench.}
    \label{fig:overview-bench}
\end{figure*}

{\bf Eliciting LLMs.} Improving LLM reasoning through elicitation has become an emerging subject \citep{tang2025polar,more2026theramind}. Techniques like In-Context Learning (ICL) \citep{koike2024outfox,yugeswardeenoo2024question}, Chain-of-Thought (CoT) \citep{godwin2024distributed,wang2024stop}, and Tree-of-Thought (ToT) \citep{zhang2024chain,zhang2024can} guide models to generate step-by-step answers and improve performance on tasks like math and logic. Yet, these methods are mostly tested on clean, synthetic datasets \citep{askarbekuly2024llm,schmidhuber2024llm,meng2026small}. Instead, we evaluate LLMs with fine-grained question structure and offer insights about LLMs eligibility on each isolated reasoning step.
\section{\bench: A Pedagogical Diagnostic Benchmark for ESTs}

\subsection{Scope and Data Collection}
\label{ssec:data}

To empirically validate our diagnostic framework, we construct \bench, a large-scale, multimodal dataset comprising \ndata questions across \nqtype distinct question types. The benchmark aggregates high-quality data from five internationally recognized English Standardized Tests (ESTs): SAT, GRE, GMAT, TOEFL, and IELTS, as summarized in Figure \ref{fig:overview-bench}, with detailed introduction and rationale of their representativeness in \textbf{Appendix \ref{app:scope}}.

\textbf{Modality.} (Details are provided in \textbf{Appendix \ref{app:modality}}). \bench is designed to reflect the complex, heterogeneous cognitive demands of real-world test-taking. The dataset spans five primary modalities: text, mathematical symbols, images, tables, and audio. This multimodal structure ensures that our evaluation extends beyond pure natural language processing to include visual parsing, spatial reasoning, and auditory comprehension.

\textbf{Question Type.} For the purpose of rigorous pedagogical diagnosis, \bench strictly focuses on \textit{objective questions} (e.g., multiple-choice and deterministic numeric entry). This constraint is necessary to provide ensure certainty from solution trajectories to answers and facilitate us to isolate exact cognitive bottlenecks without the subjective ambiguity of open-ended generation, deterministically score intermediate reasoning steps, and map each question to our cognitive framework (\S\ref{ssec:task}).

\begin{table*}[htbp]
\centering
\caption{Taxonomy of Cognitive Trajectories across EST Question Types. The traversal paths illustrate the sequential cognitive nodes required to solve the task, enabling precise diagnostic bottleneck isolation. Table \ref{tab:benchmark} provides detailed mapping from EST tasks to pedagogical categories.}
\label{tab:dag_taxonomy}
\resizebox{\textwidth}{!}{%
\begin{tabular}{@{}lllc@{}}
\toprule
\textbf{Pedagogical Category} & \textbf{Task Domain} & \textbf{Cognitive Trajectory ($C$)} & \textbf{Distractor Type} \\
\midrule
\multirow{2}{*}{\begin{tabular}[c]{@{}l@{}}\textbf{Knowledge-intensive}\\
(Lexical \& rule-based)
\end{tabular}} 
& Structural reasoning & $c_{\texttt{syntax\_parse}} \rightarrow c_{\texttt{rule\_match}} \rightarrow c_{\texttt{predict}}$ & Grammar \& syntax deficits \\
& Semantic reasoning & $c_{\texttt{semantic\_parse}} \rightarrow c_{\texttt{logical\_scope}} \rightarrow c_{\texttt{resolve}}$ & Vocabulary \& lexical gaps \\
\midrule
\multirow{2}{*}{\begin{tabular}[c]{@{}l@{}}\textbf{Reasoning-intensive} \\ (math \& multimodal)
\end{tabular}}
& Data interpretation & $c_{\texttt{analytical\_goal}} \rightarrow c_{\texttt{visual\_parse}} \rightarrow c_{\texttt{analyze}}$ & Multimodal alignment errors \\
& Numeric calculation & $c_{\texttt{model}} \rightarrow c_{\texttt{formulate\_eq}} \rightarrow c_{\texttt{compute}}$ & Symbolic \& execution errors \\
\midrule
\multirow{2}{*}{\begin{tabular}[c]{@{}l@{}}\textbf{Hybrid integration} \\(textual \& inference)
\end{tabular}}
& Evidence finding & $c_{\texttt{intent}} \rightarrow c_{\texttt{perceive\_text}} \rightarrow c_{\texttt{extract}}$ & Information retrieval failures \\
& Comparative inference & $c_{\texttt{entities}} \rightarrow c_{\texttt{constraints}} \rightarrow c_{\texttt{evaluate}}$ & Logical fallacies / hallucination \\
\bottomrule
\end{tabular}%
}
\end{table*}

\begin{figure*}[t]
    \centering
    \includegraphics[width=\textwidth]{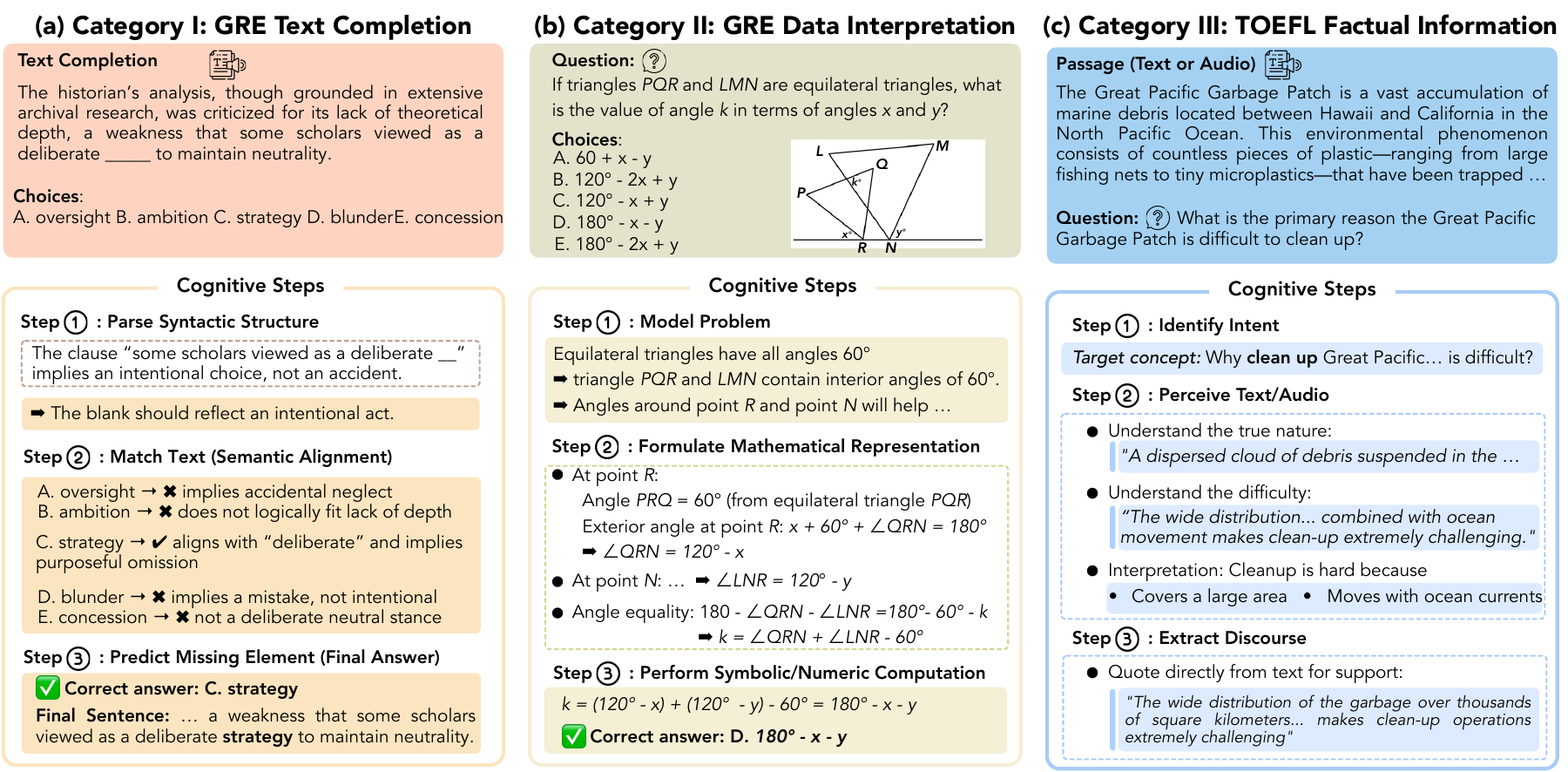}
    \caption{Illustrative cognitive trajectories for solving EST questions.}
    \label{fig:breakdown}
\end{figure*}

\textbf{Source and Quality Control.} (Details are provided in \textbf{Appendix \ref{app:source} and \ref{app:copyright}}). All data is curated from official preparation guides, released practice exams, and vetted open-access educational platforms \citep{gmac2025bundle,toefltestprep2023,woldoff2024gre,appelrouth2017preparing,josue2023educational,pereira2024leveraging}. To ensure strict \textbf{quality control and licensing compliance}, all test sources were manually validated for pedagogical fidelity, and proprietary multimodal elements were carefully reconstructed to mirror authentic test conditions.

\subsection{Constructing the Cognitive Trajectory: A Diagnostic Framework of EST Sub-skills}
\label{ssec:task}

To evaluate LLMs from the monolithic question answering to a pipelined diagnostic framework, we map each EST question type to a \textbf{Cognitive Trajectory}. Motivated by the principle of cognitive diagnosis models \citep{leighton2007cognitive} used in psychometrics, we posit that solving a EST question $Q$ (no matter complex or simple) requires traversing a specific graphical trajectory of underlying cognitive sub-skills, denoted as $C = \{c_1, c_2, \cdots, c_n \}$. 

Given that evidence-centered practice are commonly used across test centers  \citep{mislevy2003brief}, we correspondingly formalize the cognitive trajectory as a universal set of cognitive nodes $\mathcal{N}$: ranging from knowledge retrieval ($c_\texttt{lexical}$) to fluid logic ($c_\texttt{formulate}$) and final execution ($c_\texttt{compute}$). We route an LLM's reasoning trace through these trajectories and can subsequently pinpoint the exact cognitive bottleneck where an error occurs. As detailed in Table~\ref{tab:dag_taxonomy}, we consolidate the diverse formats of ESTs into three distinct pedagogical categories based on their reliance on knowledge retrieval versus multi-step reasoning:

\textbf{Category I: Knowledge-intensive retrieval (lexical \& structural):} EST questions in this category test the model’s internalized understanding of linguistic rules and vocabulary. A success cognitive process relies on localized pattern matching on two sub-categories:
\begin{itemize}
\item \underline{(I-1) Structural reasoning} questions require the progression of analyzing syntax, matching text rules, and predicting missing elements ($c_{\text{syntax\_parse}} \rightarrow c_{\text{rule\_match}} \rightarrow c_{\text{predict}}$). Examples include GRE's text completion.
\item \underline{(I-2) Semantic reasoning} questions involve interpreting fine-grained semantics and resolving logical equivalence ($c_{\text{semantic\_parse}} \rightarrow c_{\text{logical\_scope}} \rightarrow c_{\text{resolve}}$), as seen in GRE's sentence equivalence. Failure at these nodes typically indicates a pure knowledge gap rather than a reasoning flaw.
\end{itemize}

\textbf{Category II: Reasoning-intensive execution (multimodal \& quantitative):} These tasks require minimal external knowledge but demand rigorous, multi-step logic, spatial awareness, and mathematical execution. 
\begin{itemize}
    \item \underline{(II-1) Data interpretation} questions typically include multimodality involving tables or charts, the model must formulate an analytical goal, parse the non-textual visual input, and analyze the underlying data ($c_{\text{analytical\_goal}} \rightarrow c_{\text{visual\_parse}} \rightarrow c_{\text{analyze}}$). 
    
    \item \underline{(II-2) Numeric calculation} questions, e.g., SAT Math and GRE Quantitative, require translating natural language descriptions into formal mathematical models, formulating symbolic representations, and performing strict arithmetic computation ($c_{\text{model}} \rightarrow c_{\text{formulate\_eq}} \rightarrow c_{\text{compute}}$). Bottlenecks here frequently reveal that LLMs can correctly formulate an equation ($c_{\text{formulate\_eq}}$) but fail at pure arithmetic execution ($c_{\text{compute}}$), a critical pedagogical insight.
\end{itemize}

 \textbf{Category III: Hybrid integration (semantic extraction \& inference):} These tasks bridge knowledge retrieval and fluid logic. The model must synthesize large contexts (text or audio) and apply deductive reasoning to information that is not explicitly stated.
\begin{itemize}
    \item \underline{(III-1) Evidence finding} questions ask the model to identify the central intent, locate relevant evidence within a dense context, and extract the discourse ($c_{\text{intent}} \rightarrow c_{\text{perceive}} \rightarrow c_{\text{extract}}$). This trajectory is heavily utilized in TOEFL's and GRE's reading comprehension.
    \item \underline{(III-2) Comparative judgment} questions require identifying comparative entities, applying given constraints, and evaluating logical sufficiency ($c_{\text{entities}} \rightarrow c_{\text{constraints}} \rightarrow c_{\text{evaluate}}$), common in GMAT's critical reasoning.
\end{itemize}

\subsection{Data Annotation and Curation with Cognitive Trajectories and Distractor Rationale}

\bench restructures EST problem-solving into above specific cognitive trajectories.
To ensure high fidelity, we develop an NLP-driven annotation pipeline to deterministically construct the ground-truth cognitive trajectories and distractor rationales, which avoids using generative LLMs to prevent data contamination and hallucination. We first apply dependency parsing and rule-based syntactic mapping to official test-preparation explanations, wherein we chronologically extract the explicit cognitive nodes (e.g., visual parsing, equation formulation) required to solve each question. Furthermore, we categorize incorrect multiple-choice options into a rigorous taxonomy of cognitive traps (e.g., partial truth, execution error) utilizing classical NLP techniques such as lexical overlap analysis, negation scope detection, and named entity recognition. Following human-in-the-loop validation, this pipeline yields a pedagogically grounded benchmark where both correct reasoning paths and human-like failure modes are systematically formalized. Comprehensive details of this annotation methodology are deferred to Appendix \ref{app:annotation_pipeline}.

\section{Experiment}
\label{sec:expt}

\begin{table*}[!t]
\caption{Results on \bench. Human performance is reported as the average across five independent testers. For each method's results, we report the average over five independent runs with adjusted temperatures (0.2, 0.5, 0.7, 0.9, 1.0). Standard deviations are shown in \textbf{Table \ref{tab:main-expt-std}}.}
\small
\renewcommand{\arraystretch}{0.9}
\resizebox{\textwidth}{!}{
\centering
\begin{tabular}{l|c|ccc|ccc|ccc|ccc|ccc|ccc}
\toprule
\multirow{2.5}{*}{\bf Task} & \multirow{2.5}{*}{\bf Human} 
& \multicolumn{3}{c}{\bf GPT-4V} 
& \multicolumn{3}{c}{\bf GPT-5} 
& \multicolumn{3}{c}{\bf Claude-Sonnet-4} 
& \multicolumn{3}{c}{\bf Llama-4-Scout-17B} 
& \multicolumn{3}{c}{\bf Qwen-VL-Max} 
& \multicolumn{3}{c}{\bf Gemini-2.5} \\
\cmidrule(lr){3-5}\cmidrule(lr){6-8}\cmidrule(lr){9-11}\cmidrule(lr){12-14}\cmidrule(lr){15-17}\cmidrule(lr){18-20}
& & ICL & CoT & ToT & ICL & CoT & ToT & ICL & CoT & ToT & ICL & CoT & ToT & ICL & CoT & ToT & ICL & CoT & ToT \\
\midrule
\rowcolor{cellshade}
\multicolumn{20}{c}{\it \textbf{SAT}} \\
\midrule
\textbf{II} & 82.1 & 73.3 & 75.6 & 80.6 & 72.8 & 82.2 & 86.4 & 83.9 & 85.4 & 89.7 & 81.2 & 87.5 & 90.4 & 88.3 & 89.2 & 91.5 & 85.0 & 87.3 & 92.6 \\
\textbf{CS} & 74.0 & 68.2 & 77.4 & 84.9 & 73.9 & 82.7 & 87.2 & 55.8 & 70.2 & 66.3 & 46.4 & 61.8 & 55.7 & 75.6 & 82.9 & 78.8 & 93.7 & 94.1 & 87.2 \\
\textbf{EI} & 77.5 & 78.1 & 79.5 & 78.6 & 84.0 & 84.5 & 82.1 & 50.5 & 62.4 & 64.8 & 48.6 & 52.5 & 51.2 & 59.0 & 61.9 & 66.1 & 72.4 & 70.4 & 71.8 \\
\textbf{EC} & 89.0 & 84.7 & 89.3 & 81.3 & 93.8 & 92.2 & 84.7 & 72.0 & 74.2 & 70.1 & 64.0 & 65.4 & 63.9 & 81.3 & 77.2 & 79.3 & 93.3 & 90.6 & 92.1 \\
\textbf{AG} & 55.1 & 28.4 & 44.0 & 60.9 & 31.7 & 53.2 & 76.1 & 33.3 & 52.6 & 79.4 & 30.0 & 46.7 & 68.3 & 35.8 & 50.1 & 81.6 & 34.2 & 52.7 & 82.4 \\
\textbf{DA} & 77.9 & 56.7 & 70.9 & 85.8 & 60.3 & 78.2 & 90.7 & 58.2 & 71.4 & 90.1 & 51.1 & 60.2 & 87.3 & 54.6 & 67.2 & 89.2 & 53.9 & 69.7 & 88.0 \\
\textbf{GT} & 63.0 & 66.7 & 64.7 & 67.3 & 73.9 & 70.5 & 71.6 & 49.7 & 50.8 & 47.2 & 41.2 & 44.8 & 38.1 & 33.3 & 30.4 & 32.5 & 58.8 & 59.0 & 56.2 \\
\midrule
\rowcolor{cellshade}
\multicolumn{20}{c}{\it \textbf{GRE}} \\
\midrule
\textbf{TC }& 76.2 & 72.6 & 77.4 & 83.1 & 68.5 & 73.4 & 82.1 & 69.4 & 75.5 & 72.4 & 53.5 & 61.0 & 64.8 & 67.7 & 73.1 & 78.3 & 68.5 & 80.2 & 82.4 \\
\textbf{SE} & 81.5 & 78.9 & 81.0 & 79.8 & 87.7 & 86.5 & 87.2 & 85.5 & 82.1 & 83.5 & 66.0 & 67.5 & 63.2 & 71.8 & 73.2 & 71.6 & 77.6 & 74.8 & 75.9 \\
\textbf{RC} & 70.2 & 67.1 & 77.8 & 86.1 & 83.6 & 87.1 & 81.5 & 61.9 & 69.3 & 76.0 & 46.3 & 54.2 & 73.2 & 70.6 & 76.2 & 80.1 & 56.9 & 73.2 & 78.6 \\
\textbf{QC} & 68.1 & 55.3 & 57.3 & 51.3 & 82.0 & 84.1 & 83.8 & 41.3 & 48.2 & 44.6 & 51.3 & 56.0 & 42.7 & 54.7 & 50.3 & 45.7 & 48.0 & 58.4 & 53.1 \\
\textbf{NE} & 73.7 & 32.7 & 38.0 & 52.7 & 28.7 & 33.9 & 48.2 & 17.3 & 25.0 & 37.2 & 23.3 & 30.1 & 44.5 & 29.3 & 28.1 & 40.8 & 26.0 & 33.0 & 30.2 \\
\textbf{DI} & 55.5 & 52.0 & 56.0 & 73.3 & 32.7 & 36.5 & 63.2 & 21.3 & 25.7 & 50.1 & 40.0 & 41.2 & 65.1 & 38.7 & 40.5 & 61.7 & 48.0 & 47.2 & 67.1 \\
\midrule
\rowcolor{cellshade}
\multicolumn{20}{c}{\it \textbf{GMAT}} \\
\midrule
\textbf{CR} & 66.2 & 62.3 & 77.9 & 72.5 & 57.4 & 70.1 & 71.4 & 55.7 & 79.5 & 74.8 & 65.6 & 69.2 & 71.3 & 57.4 & 75.6 & 70.2 & 56.1 & 74.4 & 72.7 \\
\textbf{RC} & 82.1 & 79.2 & 88.7 & 91.4 & 65.2 & 71.4 & 75.6 & 63.5 & 81.1 & 86.2 & 47.3 & 74.5 & 70.3 & 68.6 & 74.4 & 76.8 & 59.1 & 75.0 & 77.4 \\
\textbf{PS} & 73.7 & 24.0 & 34.3 & 41.2 & 26.0 & 31.1 & 54.2 & 19.1 & 24.5 & 27.2 & 18.6 & 22.5 & 35.0 & 22.1 & 25.6 & 33.7 & 25.0 & 28.3 & 38.4 \\
\textbf{DS} & 52.0 & 14.5 & 26.8 & 24.5 & 13.5 & 32.4 & 40.8 & 12.0 & 16.0 & 19.2 & 13.8 & 14.5 & 20.1 & 14.8 & 21.0 & 23.6 & 9.0 & 13.5 & 22.0 \\
\textbf{IR} & 59.2 & 11.2 & 13.8 & 22.1 & 11.8 & 16.0 & 20.3 & 8.8 & 15.0 & 17.4 & 3.2 & 16.2 & 18.0 & 10.0 & 11.2 & 18.7 & 12.1 & 14.4 & 20.5 \\
\midrule
\rowcolor{cellshade}
\multicolumn{20}{c}{\it \textbf{TOEFL}} \\
\midrule
\textbf{FI }& 86.5 & 82.3 & 86.3 & 74.2 & 85.5 & 93.2 & 70.5 & 76.6 & 83.9 & 82.0 & 65.3 & 68.8 & 65.7 & 73.2 & 70.5 & 75.1 & 73.5 & 84.1 & 86.3 \\
\textbf{IR} & 74.1 & 63.4 & 85.3 & 87.7 & 79.3 & 84.2 & 85.0 & 55.9 & 59.2 & 63.0 & 46.0 & 62.2 & 58.3 & 73.5 & 74.0 & 75.2 & 79.0 & 81.0 & 82.6 \\
\textbf{TS} & 85.0 & 83.9 & 86.1 & 84.8 & 83.9 & 84.0 & 81.7 & 83.5 & 85.0 & 82.4 & 74.2 & 75.8 & 73.0 & 73.5 & 75.5 & 76.2 & 66.1 & 67.0 & 66.8 \\
\textbf{FI} & 93.1 & 93.7 & 95.7 & 97.7 & 94.0 & 93.2 & 98.5 & 80.7 & 86.5 & 82.5 & 67.7 & 69.2 & 76.6 & 74.7 & 70.8 & 76.3 & 81.3 & 92.5 & 89.7 \\
\textbf{IF} & 70.1 & 62.0 & 64.7 & 67.3 & 81.3 & 88.4 & 90.8 & 70.7 & 82.0 & 79.1 & 55.3 & 58.8 & 61.2 & 53.3 & 62.4 & 55.8 & 68.0 & 72.8 & 80.2 \\
\midrule
\rowcolor{cellshade}
\multicolumn{20}{c}{\it \textbf{IELTS}} \\
\midrule
\textbf{II} & 82.0 & 79.1 & 84.8 & 82.8 & 81.1 & 86.0 & 88.4 & 79.1 & 82.0 & 79.5 & 75.7 & 74.5 & 71.3 & 73.0 & 76.2 & 74.1 & 83.1 & 84.2 & 86.0 \\
\textbf{MS} & 93.6 & 83.7 & 85.1 & 81.3 & 81.7 & 83.0 & 83.7 & 73.1 & 81.0 & 83.2 & 66.8 & 74.0 & 76.0 & 69.2 & 71.2 & 73.7 & 75.5 & 82.5 & 80.4 \\
\textbf{CP} & 71.8 & 66.0 & 67.2 & 72.1 & 83.1 & 82.4 & 84.0 & 71.8 & 84.4 & 85.5 & 58.4 & 73.1 & 75.6 & 73.5 & 72.4 & 76.7 & 82.1 & 81.0 & 83.9 \\
\textbf{IM} & 86.1 & 83.7 & 84.8 & 88.3 & 90.6 & 91.5 & 92.8 & 74.0 & 76.0 & 75.1 & 64.2 & 66.4 & 68.3 & 73.1 & 72.0 & 74.8 & 83.7 & 89.2 & 91.6 \\
\textbf{CL} & 88.3 & 80.5 & 84.6 & 83.1 & 83.6 & 91.0 & 90.4 & 72.5 & 74.8 & 73.0 & 41.3 & 61.0 & 66.7 & 58.2 & 64.4 & 67.3 & 76.1 & 82.0 & 88.5 \\
\textbf{SA} & 85.1 & 83.0 & 86.4 & 84.7 & 83.0 & 85.1 & 84.0 & 73.9 & 77.0 & 75.0 & 66.2 & 70.2 & 67.6 & 73.3 & 76.4 & 74.7 & 82.1 & 84.9 & 83.2 \\
\bottomrule
\end{tabular}}
\label{tab:main-expt}
\end{table*}

\bench aims to empirically answer several research questions:
{\bf RQ$_1$:} How do state-of-the-art LLMs perform on EST questions?
{\bf RQ$_2$:} How effectively do LLMs execute cognitive reasoning steps within the cognitive framework? {\bf RQ$_3$:} How robust are LLMs to distractor options? {\bf RQ$_4$:} How to improve LLMs' pedagogical reasoning based on previous experimental findings?

{\bf LLMs.} Given the multimodal nature of \bench, we evaluate several industry-leading Multimodal LLMs, including GPT-5, GPT-4V, Claude-Sonnet-4, Llama-4-Scout-17B, Qwen-VL-Max, and Gemini-2.5.
We adopt OpenAI's Whisper \citep{andreyev2025quantization,graham2024evaluating} to transcribe audio data within listening tasks in TOEFL and IELTS.

{\bf Human Tester.} To demonstrate LLM performance alongside humans, we included five student testers with their consent to share the data, wherein each tester had recently prepared for and participated in at least one of these tests.

{\bf Prompting Strategy.} We evaluate three popular prompting methods:
{\bf (1) In-Context Learning (ICL)}: Besides basic instructions to describe the question type, the prompt also includes several (we select five) examples to offer LLMs the solution style.
{\bf (2) Chain-of-Thought (CoT)} \citep{bi2025forestofthoughtscalingtesttimecompute,zhang2024chain}: The prompt encourages the model to generate intermediate reasoning steps before generating final answers.
{\bf (3) Tree-of-Thought (ToT)} \citep{long2023large,yao2023tree}: An advanced strategy that guides LLMs to explore multiple reasoning paths and select the most plausible one. Prompts are shown in \textbf{Appendix \ref{app:prompt}}.


\subsection{Evaluating LLMs Performance (RQ$_1$)}
\label{ssec:expt-main}

\underline{{\bf Problem-solving abilities.}} Table~\ref{tab:main-expt} presents the performance of various LLMs on \bench. Despite extensive pretraining on large-scale English corpora, these models exhibit substantial variability across different EST tasks, even within similar domains and modalities. For instance, in linguistic tasks such as GRE Expression of Ideas (EI) and English Conventions (EC), GPT-4V achieves 79.5\% and 89.3\% accuracy, respectively, revealing its inconsistent ability to handle fine-grained distinctions in grammar, style, and logical flow. Similarly, LLMs do not always outperform human testers despite their advanced prompting methods (e.g., COT or TOT). Those observations suggest that LLMs often struggle with the contextual sensitivity required for generalizing to diverse test problems. We provide more results in \textbf{Appendix \ref{app:add-expt-rq1}}.

 \begin{figure*}[h]
    \centering
    \includegraphics[width=0.98\textwidth]{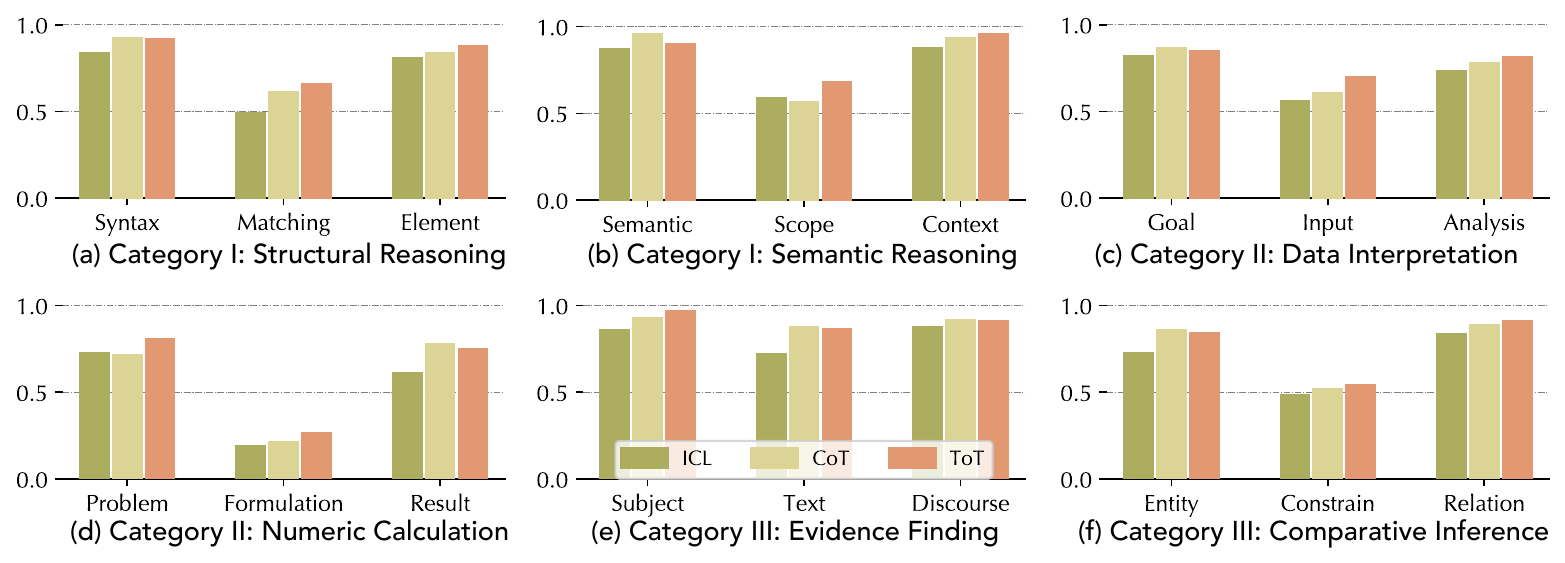}
    \caption{Detailed stepwise analysis on GPT-5 on each cognitive step.}
    \label{fig:expt-breakdown-gpt4o}
\end{figure*}

\underline{{\bf Influence by modality.}}
The limitations of LLMs become more obvious when complex modalities are involved, such as GMAT Integrated Reasoning (IR) and GRE Data Interpretation (DI).

\begin{theobox}
{\bf Case Study.} (GMAT -- Integrated Reasoning): You are given (i) A {\bf table} showing sales data by region and quarter (e.g., North America). (ii) A {\bf text passage} describing factors that influenced sales in different regions (e.g., {\it "A new competitor entered the European market in Q2..."}).

\vspace{3pt}
{\bf Question:}
\textit{Which region experienced the largest relative revenue drop between Q1 and Q2?}

\vspace{3pt}
{\bf Challenges for LLMs:}
{\bf 1. Mapping text to table:}
Claude and GPT-5 fail to connect the textual clue ({\it "new competitor in Europe"}) with the relevant table entry ({\it Europe, Q2 revenue}). 
{\bf 2. Reasoning with partial information:} Gemini overlooks the hint about the {\it competitor’s impact} and fails to compare percentage drops across regions, missing the correct answer.
\end{theobox}
The challenge is twofold: first, models must align disparate representations (e.g., mapping textual queries to tabular structures); second, they must reason over incomplete or distributed evidence, a skill that current architectures and training regimes are not fully optimized for. 

These multimodal failures suggest that achieving human-level performance on ESTs requires more than language modeling proficiency; it demands integrated reasoning capabilities that span visual, symbolic, and logical modalities. Together, these observations highlight the inherent difficulty of EST-style questions and the under-preparedness of even the strongest LLMs to serve as reliable tutors for real-world educational settings. We provide more studies and failure modes in \textbf{Appendix \ref{app:rq1-modality}}.

\ul{\textbf{More analysis.}} Due to space constraints, we defer additional analyses and discussions to the appendix: \textbf{(1) App. \ref{app:expt-human}} quantifies how LLMs align with human testers and includes statistical tests demonstrating the significance of behavioral similarities between LLMs and human testers. \textbf{(2) App. \ref{app:expt-prompt-complexity}} examines the impact of different prompting strategies (ICL, CoT, and ToT) with supporting case studies. \textbf{(3) App. \ref{app:expt-difficulty}} analyzes the influence of question difficulty. \textbf{(4) App. \ref{app:expt-time}} quantifies the answering time of LLMs.

 \begin{figure*}[h]
    \centering
    \includegraphics[width=\textwidth]{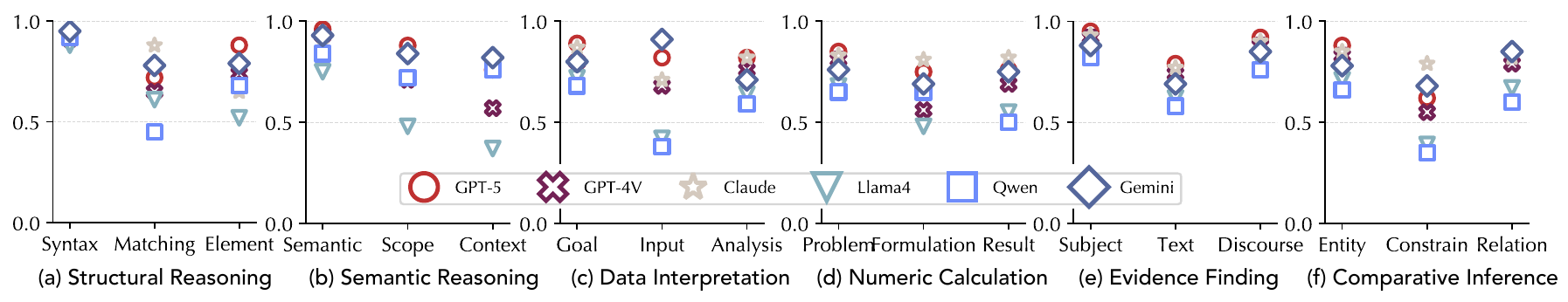}
    \caption{Analysis LLMs performance to distractor options for each cognitive step.}
    \label{fig:expt-distractor}
\end{figure*}

\subsection{Isolated Measurement of Each Cognitive Step (RQ$_2$)}
\label{ssec:expt-breakdown}

Following our formalized cognitive trajectories (\S\ref{ssec:task}), we evaluate the step-by-step reasoning capabilities of LLMs using ground-truth trajectories extracted via our deterministic annotation pipeline (\textbf{Appendix \ref{app:annotation_pipeline}}). Because each cognitive node exhibits distinct properties. For instance, syntactic parsing requires exact matching, whereas numeric computation necessitates error tolerance. We thus deploy a tailored suite of node-level metrics (e.g., IoU, BERTScore, and Normalized RMSE) to address the specific diagnostic requirements of each reasoning stage with full metric definitions articulated in \textbf{Appendix \ref{app:breakdown_metrics}}. Figure \ref{fig:expt-breakdown-gpt4o} and \textbf{Appendix \ref{app:add-expt-rq2}} present the performance breakdown across all cognitive steps. From this stepwise analysis, we derive the following key observations:

\ul{\bf LLMs are generally strong formulators but occasionally weak reasoners.}
Overall, we observe that LLMs consistently excel in the initial step across all tasks, achieving up to 97\% accuracy. These early steps demonstrate the models’ strong capability to interpret and structure EST problems appropriately, such as task identification, problem formulation, or topic modeling. However, performance significantly declines in subsequent reasoning steps, which vary across tasks. This decline is particularly evident in tasks that demand causality inference or evidence synthesis. 

\begin{theobox}
{\bf Case Study (GMAT -- Critical Reasoning):} "{\it ... cities with more EV charging stations tend to have lower levels of air pollution. As a result, the city of Greentown has decided to install a large number of EV charging stations ...}"

\vspace{3pt}
{\bf LLM Reasoning:} GPT-5 incorrectly state {\it "The city of Greentown currently has a very small number of EVs in use"} due to being distracted by the phrase {\it "the current number of EVs} that does not directly relevant to causal logic.



\end{theobox}

These results suggest that LLMs remain limited and unstable in executing complex reasoning chains, which is an essential requirement for robust educational support.

\ul{\bf Complex logic has more impact on LLM screening than long context.}
We find that context length alone does not impede LLMs to locate relevant information. Instead, reasoning complexity plays a greater role in determining success or failure. Models can navigate long inputs effectively if the task only requires surface-level matching, but they often fail when logical integration across multiple sentences is required. These results suggest that long context alone is not a major barrier for modern LLMs. However, once the task requires multi-hop reasoning or integrating dispersed evidence, even top-performing models struggle.

\ul{\textbf{More analysis and case studies.}} Due to space constraints, additional findings and case studies are provided in \textbf{App. \ref{app:add-expt-rq2}} and \textbf{\ref{app:more-case-study}}.

\subsection{LLM Performance on Distractors (RQ$_3$)}

After RQ$_2$ evaluates LLMs performance on ground-truth answers, we diagnose LLMs regarding their robustness on distractor options. For each task, we prompt the LLMs with the incorrect distractor options and a rationale to ask them to accept or reject each distractor options. We report the overall accuracy of selection following the 3-step cognitive trajectory. If the model hallucinates a selection (e.g., erroneously accept a distractor), it receives a score of zero for that instance.

\ul{\bf Independence of distractor susceptibility and generative robustness}
Comparing our results from RQ$_2$ (performance on ground-truth results) with RQ$_3$ (robustness against distractor options), we observe a clear  independence between generative problem-solving and discriminative robustness. Models like GPT-5 and Claude demonstrate notably degrades when explicitly prompted to evaluate and reject carefully engineered distractor rationales. For instance, in Category I (Semantic Reasoning), although models exhibit strong generative capabilities, introducing distractors that utilize partial semantic truths causes severe performance drops. This disentanglement implies that high accuracy on standard benchmarks (RQ$_2$) does not guarantee pedagogical robustness. As LLMs work as plausible reasoners rather than logical discriminators, highly susceptible behaviors to the same cognitive traps may misguide human test-takers.

\ul{\textbf{Failure mode study.}} Due to space constraints, we defer the failure model analysis to \textbf{App. \ref{app:failure_mode_analysis}}.

\subsection{Improvement and Mitigation (RQ$_4$)}
\label{sec:mitigation}

\begin{table}[t]
\caption{Effect of targeted mitigation strategies across representative \bench tasks. Scores reflect baseline Chain-of-Thought (CoT) versus our bottleneck-specific Mitigated (Mit.) prompting.}
\label{tab:mitigation}
\centering
\resizebox{\columnwidth}{!}{
\begin{tabular}{l cc cc cc}
\toprule
\multirow{2}{*}{\textbf{Task \& Mitigation}} & \multicolumn{2}{c}{\textbf{GPT-4V}} & \multicolumn{2}{c}{\textbf{GPT-5}} & \multicolumn{2}{c}{\textbf{Claude-S-4}}\\
\cmidrule(lr){2-3} \cmidrule(lr){4-5} \cmidrule(lr){6-7}
& Base & +Mit. & Base & +Mit. & Base & +Mit. \\
\midrule
\textbf{GRE-RC} \textit{\footnotesize + Evidence-Anchored} & 77.8 & \textbf{93.5} & 87.1 & \textbf{90.4} & 69.3 & \textbf{84.1} \\
\textbf{GRE-TC} \textit{\footnotesize + Syntax-First} & 79.5 & \textbf{84.2} & 73.4 & \textbf{78.1} & 61.0 & \textbf{76.7} \\
\textbf{GMAT-IR} \textit{\footnotesize + Table-Alignment} & 13.8 & \textbf{59.7} & 16.0 & \textbf{71.9} & 15.0 & \textbf{69.4} \\
\bottomrule
\end{tabular}
}
\end{table}

We design a set of task-specific elicitation strategies to fill in the performance gap identified from RQ$_1$-RQ$_3$. In cognitive trajectory, we enforce stepwise constraints before a model is permitted to generate its final evaluation. For instance, we introduce an \textbf{evidence-anchored CoT} in GRE reading comprehension (RC) that mandates verbatim text extraction prior to high-level interpretation, a \textbf{syntax-first CoT} in GRE-TC forces the explicit identification of grammatical roles (e.g., contrast markers) to stop models from conflating semantic plausibility with syntactic validity, and a \textbf{table-alignment Constraint} in GMAT-IR that requires explicit textual-to-tabular mapping instructions before numerical operations begin. As shown in Table \ref{tab:mitigation}, isolating and promoting these cognitive steps yield consistent performance gains across all frontier models, showing that our pedagogical framework can guide actionable mitigation.

We defer\textbf{ (1)} comprehensive mitigation results to \textbf{App. \ref{app:expt-more-rq4}}, \textbf{(2)} additional mitigation strategies and discussions to \textbf{App. \ref{app:add-miti}}, and \textbf{(3)} insights to advance LLMs in pedagogical scenarios to \textbf{App. \ref{app:advance-llm}}.

\section{Conclusion}

This work explores LLM performance on ESTs with the introduction of \bench. We move beyond surface-level accuracy metrics to model problem-solving as a structured cognitive trajectory. Through extensive experiments, we observe that current LLMs possess the linguistic fluency to conceptualize problems and execute computations, while their generative bias toward plausible rationalization making them highly susceptible to distractor traps during intermediate constraint mapping. We further introduce targeted mitigation strategies to improve LLM robustness in pedagogical settings. Ultimately, this research provides a rigorous diagnostic perspective and an actionable way toward more pedagogically sound AI systems.
\bibliography{reference}

\appendix
\section{Comprehensive Details of \bench Data Collection}
\label{app:data_collect}

This section provides extended details on the scope, modalities, sourcing, and quality control processes used to construct the \bench dataset, expanding upon the summary provided in Section \ref{ssec:data}.

\subsection{Scope and Cognitive Targets of Selected ESTs}
\label{app:scope}

\bench aggregates questions from five major standardized tests, chosen for their distinct cognitive requirements and global educational impact:

\textbf{(i) SAT:} Widely used for US undergraduate admissions, the SAT provides a baseline for evaluating foundational verbal and mathematical reasoning among high school graduates.

\textbf{(ii) GRE:} Designed for graduate admissions, the GRE tests advanced cognitive constructs, including highly complex semantic reasoning (e.g., Sentence Equivalence) and fluid quantitative logic.

\textbf{(iii) GMAT:} Functioning as the primary gatekeeper for business schools, the GMAT emphasizes executive-level logic, including Integrated Reasoning (synthesizing multimodal data) and Data Sufficiency (evaluating logical constraints without executing the math).

\textbf{(iv) TOEFL \& IELTS:} These exams assess non-native English proficiency. In \bench, they serve as the primary benchmark for evaluating an LLM's capacity for long-context auditory comprehension (audio modalities) and pragmatic textual inference.

\subsection{Multimodal Distribution and Complexity}
\label{app:modality}

Unlike traditional NLP benchmarks that rely solely on text, \bench enforces a heterogeneous multimodal environment. The modalities incorporated include:

\textbf{Text (T) \& Audio (A):} Primarily featured in TOEFL and IELTS, evaluating the model's ability to extract key discourse from dense textual passages or spoken academic lectures.

\textbf{Math Symbols (S):} Featured heavily in the SAT, GRE, and GMAT, requiring models to parse LaTeX-formatted equations and perform symbolic manipulation.

\textbf{Images (I) \& Tables (Tb):} Critical for the GRE Data Interpretation and GMAT Integrated Reasoning sections. These questions require the model to perform \textit{visual parsing} (e.g., reading a scatterplot or cross-referencing multi-column tables) before executing any mathematical reasoning.

\begin{table*}[!t]
\caption{Question types, their descriptions, number of instances, involved modalities, and their pedagogical categories (defined in \S\ref{ssec:task}). Categories: \textbf{I} -- Knowledge-Intensive Retrieval; \textbf{II} -- Reasoning-Intensive Execution; \textbf{III} -- Hybrid Integration. Modalities: "T"--text, "S"--math symbol, "I"--image, "Tb"--tabular data, "A"--audio. Concrete examples are shown in Appendix \ref{app:q-type}.}
\renewcommand{\arraystretch}{0.95}
\small
\resizebox{\textwidth}{!}{
\centering
\begin{tabular}{llllll}
\toprule
{\bf Section} & {\bf Question Type (Abbreviation)} & {\bf Description} & {\bf Num} & {\bf Modality} & {\bf Category} \\
\lightmidrule
\multicolumn{6}{c}{\bf SAT} \\
\lightmidrule
Reading \& & Information and Ideas (II) & Assess comprehension, reasoning, and inference skills & 180 & T & III \\
Writing  & Craft and Structure (CS) & Test vocabulary and how authors structure their writing & 636 & T & I \\
& Expression of Ideas (EI) & Test the logical flow and effectiveness of writing & 210 & T & I \\
&  English Conventions (EC) & Focus on grammar, punctuation, and sentence structure & 150 & T & I \\
\lightmidrule
Math & Algebra (AG) & Test numeric equations, functions, and inequalities & 243 & T,S & II \\
& Data Analysis  (DA) & Interpret ratios, percentages, probabilities, and graphs & 141 & T,S & II \\
& Geometry \& Trigonometry (GT) & Analyze angles, circles, areas, and trigonometric functions & 153 & T,S & II \\
\lightmidrule
\multicolumn{6}{c}{\bf GRE} \\
\lightmidrule
Verbal & Text Completion (TC) & Fill in blank(s) (one/two/three) within a short passage & 620 & T & I \\
& Sentence Equivalence (SE) & Choose two words with the same meaning & 620 & T & I \\
& Reading Comprehension (RC) & Answer questions based on a passage & 562 & T & III \\
\lightmidrule
Quantitative & Quant Comparison (QC) & Compare two quantities and select their relationship  & 150 & T,S & II \\
& Numeric Entry (NE) & Type the exact numerical answer & 150 & T,S & II \\
& Data Interpretation (DI) & Multi-choice questions from graphs, tables, or charts & 150 & T,S,I,Tb & II \\
\lightmidrule
\multicolumn{6}{c}{\bf GMAT} \\
\lightmidrule
Verbal & Critical Reasoning (CR) & Analyze and evaluate an argument & 244 & T & III \\
& Reading Comprehension (RC) & Answer questions based on a passage & 408 & T & III \\
\lightmidrule
Quantitative & Problem Solving (PS) & Algebra, arithmetic, numerical, and statistical problems & 408 & T,S & II \\
\lightmidrule
Data Insights  & Data Sufficiency (DS) & Decide if a statement is sufficient to answer a question & 400 & T,S & II \\
& Integrated Reasoning (IR)& Analyze tables, graphs, charts, or multiple sources  & 340 & T,S,I,Tb & II \\
 \lightmidrule
\multicolumn{6}{c}{\bf TOEFL} \\
 \lightmidrule
Reading & Factual Information (FI) & Identify facts in (or not in) the passage & 620 & T & III \\
& Inference \& Reference (IR) & Infer information/word meaning/pronoun in context & 415 & T & III \\
& Text \& Sentence (TS) & Insert texts, simplify a sentence, summarize a passage & 310 & T,Tb & I, III \\
 \lightmidrule
Listening & Factual Information (FI) & Identify facts in (or not in) the lecture/conversation & 300 & A,T & III \\
& Inference (IF) & Understand tone/intention/opinion/relationship of ideas & 150 & A,T & III \\
\lightmidrule
\multicolumn{6}{c}{\bf IELTS} \\
 \lightmidrule
 Reading & Identifying Information (II) & Identify correctness of statement or author's opinion & 296 & T & III \\
& Matching Sentence  (MS) & Match head, opinion, or sentence endings & 208 & T & III \\
& Completion (CP) & Complete sentence/summary/note/table/diagram label & 592 & T,I,Tb & I, III \\
\lightmidrule
 Listening & Identification \& Matching (IM) & Determine correct answers from the  audio  & 520 & A,T & III \\
& Completion \& Labeling  (CL) & Complete a sentence or visual with words from the audio & 1048 & A,T,I,Tb & I, III \\
& Short Answer (SA) & Answer briefly using words from the recording & 352 & T,I,Tb & III \\
\bottomrule
\end{tabular}}
\label{tab:benchmark}
\end{table*}

\subsection{Data Source}
\label{app:source}

To ensure the dataset accurately reflects the difficulty and style of modern ESTs, we sourced raw materials from a blend of highly vetted public and official resources:

\textbf{(i) Official Preparation Materials:} We extracted benchmark examples from official guides and released practice exams \citep{gmac2025bundle,toefltestprep2023,woldoff2024gre,college_board_digital_sat,hatch2023gmat}.

\textbf{(ii) Public Educational Platforms \& Practice Books:} Additional diversity in question format and difficulty was achieved by sourcing from recognized test-prep literature \citep{appelrouth2017preparing,gruber2011gruber,woldoff2015gre} and open-access educational platforms \citep{josue2023educational,pereira2024leveraging,satquestions2023,sat_suite_question_bank,gmat_club,ieltsup2023}.

\subsection{Question Quality, Copyright and Ethical Assurance}
\label{app:copyright}

To ensure the integrity of our benchmark, we adopted a rigorous process for validating both the quality of collected questions and the copyright compliance of the sources.

\textbf{Question Quality.} All questions included in \bench were sourced from publicly released or openly accessible educational materials. We verified each item for (1) correctness, by cross-checking answer keys or explanatory notes provided by the original source; (2) clarity, ensuring that wording, figures, and formatting matched the original intent without ambiguity; and (3) authenticity, by aligning the question style and content with the design principles of the corresponding standardized test (SAT, GRE, GMAT, TOEFL, or IELTS).

\paragraph{Alignment Validation Process.} 
In addition to verifying correctness and clarity, we systematically validated the alignment of our questions with official exam specifications. For each question type, we performed the following validation steps:
\begin{enumerate}
    \item \textbf{Format validation:} We cross-referenced the question structure, answer format, and presentation style against official sample materials to ensure consistency. As an example, SAT Math Algebra questions in our benchmark follow the same five-option multiple-choice format and use identical equation presentation conventions as those in College Board released items.
    
    \item \textbf{Skill mapping:} We verified that each question assesses the specific cognitive skill as defined in official test frameworks. For instance, TOEFL Reading ``Factual Information'' questions in ESTBOOK are designed to test the ability to identify explicitly stated details, which matches the skill definition provided by ETS. Similarly, ``Inference'' questions require test-takers to draw logical conclusions that go beyond the literal text, consistent with official specifications.
    
    \item \textbf{Difficulty calibration:} Although we do not have access to proprietary difficulty ratings used by test administrators, we ensured that our question collection spans the full difficulty range found in official preparation materials. This includes both entry-level items suitable for beginners and advanced problems that challenge high-performing test-takers, providing representative coverage across each exam's difficulty spectrum.
\end{enumerate}
Through this systematic validation process, combined with our reliance on officially affiliated sources, we ensure that ESTBOOK authentically captures the reasoning demands and structural characteristics of real standardized tests.

\textbf{Copyright and Ethical Usage.} To comply with licensing and intellectual property requirements, we restricted data collection to (1) officially released practice tests and preparation guides distributed for public use, and (2) open-access educational platforms and community-contributed question banks that explicitly allow free access for study and research purposes. No proprietary or paywalled materials were included. For multimodal reconstructions, all recreated content is original and designed solely to approximate the test-taking context without replicating copyrighted assets. This approach ensures that \bench respects copyright protections while still providing representative and high-quality benchmark content.

\section{Rigorous Annotation Pipeline for Cognitive Trajectories and Distractor Rationales}
\label{app:annotation_pipeline}

To ensure the high fidelity and reproducibility of the \bench dataset, we implemented a deterministic, NLP-driven annotation pipeline. Our priority is to avoid relying on generative LLMs for the creation of intermediate reasoning steps or distractor rationales, as doing so risks introducing LLM-inherent biases, hallucinations, or data contamination into the evaluation benchmark. Instead, our methodology strictly relies on parsing established ground-truth text using classical NLP techniques.

The annotation process synthesizes official test-preparation explanations (e.g., from ETS or GMAC, which provide explicit rationales for both correct and incorrect options) with deterministic NLP frameworks. The pipeline consists of three primary phases:

\subsection*{Phase 1: Stepwise Extraction via Syntactic Parsing}
Standardized test explanations typically present reasoning in a sequential, textual format. To map these raw textual explanations into our formalized cognitive trajectory ($C = \{c_1, c_2, \dots, c_n\}$), we employ a deterministic information extraction (IE) pipeline built on dependency parsing and semantic role labeling (SRL).

\begin{enumerate}
\item \textbf{Sentence Segmentation and Dependency Parsing:} We process the official explanation for the correct answer using a standard dependency parser (e.g., spaCy). By analyzing syntactic trees, we isolate sequential imperative or action-oriented clauses (e.g., \textit{"First, calculate the volume..."} $\rightarrow$ [VERB: calculate, DOBJ: volume]).
\item \textbf{Rule-Based Node Mapping:} We utilize a predefined lexicon of cognitive triggers to map the parsed clauses to specific cognitive steps. For example: Lexical triggers such as \textit{"according to the passage,"} or \textit{"line 15 states"} are deterministically mapped to $c_{\text{perceive\_text}}$ (Evidence Extraction). Mathematical operators or numbers extracted via regular expressions (Regex) alongside terms like \textit{"let x be"} or \textit{"formula"} are mapped to $c_{\text{formulate\_eq}}$ (Symbolic Mapping).

\item \textbf{Sequential Alignment:} The parsed nodes are chronologically ordered to establish the ground-truth traversal path for the question, ensuring each step logically precedes the next without LLM intervention.
\end{enumerate}

\subsection*{Phase 2: Distractor Rationale Annotation and Taxonomy Mapping}
Evaluating a model's pedagogical utility requires understanding \textit{why} it selects an incorrect option. Official test materials often include explanations for distractors (e.g., \textit{"Option B is incorrect because it misinterprets the data in Figure 1"}). To formalize this, we developed a rigid taxonomy of cognitive traps and utilized traditional NLP to annotate them.

We define four primary distractor categories for ESTs: \textbf{Type I: Direct Contradiction} (The option states the exact opposite of the text/data).
 \textbf{Type II: Out of Scope} (The option relies on information not provided in the prompt).
\textbf{Type III: Partial Truth / Faulty Premise} (The option accurately quotes the text but misapplies it to the specific question intent).
\textbf{Type IV: Execution / Calculation Error} (The option represents a common arithmetic mistake or formula misapplication).

To automate the annotation of these distractors across \ndata questions, we apply the following non-generative NLP techniques:
\begin{enumerate}
\item \textbf{Lexical Overlap and TF-IDF:} We calculate the lexical overlap between the distractor text, the raw explanation, and the source passage. Distractors exhibiting high lexical overlap with the passage but flagged as incorrect in the raw explanation are commonly classified as \textit{Type III: Partial Truth}, representing a deliberate cognitive trap designed to penalize superficial pattern-matching.
\item \textbf{Negation Scope Detection:} Using classical syntactic parsing, we detect negation modifiers (e.g., \textit{not, never, lacks}) within the raw explanation of the wrong answer. If the explanation explicitly points out a negation mismatch between the distractor and the text, it is tagged as \textit{Type I: Direct Contradiction}.
\item \textbf{Entity and Variable Mismatch:} For quantitative reasoning, we use Named Entity Recognition (NER) and Regex to extract the mathematical entities in the distractor. If the distractor matches the output of a partially completed equation (extracted from the cognitive step $c_{\text{formulate\_eq}}$ but stopping before $c_{\text{compute}}$), it is deterministically tagged as \textit{Type IV: Calculation Error}.
\end{enumerate}

\subsection*{Phase 3: Human-in-the-Loop Validation}
To ensure the absolute integrity of the benchmark, a random subset of the NLP-annotated trajectories and distractor rationales was subjected to expert review. Domain experts (educators familiar with EST design) evaluated the alignment between the deterministic NLP tags and the pedagogical intent of the question, achieving a high inter-rater reliability (Cohen’s $\kappa > 0.85$). We thus confirm that our non-AI methodology successfully and rigorously extracts the cognitive framework required for pedagogical evaluation.

\section{Complementary Information of Question Types}
\label{app:q-type}

This appendix provides a brief description of each question type covered across SAT, GRE, GMAT, TOEFL, and IELTS, along with a concrete example.  In the examples below, long passages are truncated with “\dots” for brevity.

\noindent\textbf{SAT Reading \& Information and Ideas (II).}  
Tests comprehension of written passages across diverse subjects. Students must identify main ideas, understand explicit details, make logical inferences, and draw evidence-based conclusions. Requires distinguishing between stated facts and implied meanings while recognizing the author's purpose and how ideas relate to one another. Success depends on analytical reading rather than simply locating isolated facts.

\begin{tcolorbox}[
    colback=white,
    colframe=black,
    width=\columnwidth,
    boxrule=0.5pt,
    arc=3pt,
    left=5pt,
    right=5pt,
    boxsep=0pt,
    top=3pt,
    bottom=3pt,
]
Passage: The industrial revolution transformed societies by shifting labor from farms to factories, fundamentally altering social structures and economic relationships. As rural populations migrated to urban centers seeking employment, traditional family units were disrupted...

Question: According to the passage, what was one major social change brought about by the industrial revolution?\\
A. The development of new agricultural techniques\\
B. The disruption of traditional family structures\\
C. The elimination of class divisions in society\\
D. The reduction in opportunities for social mobility
\end{tcolorbox}

\noindent\textbf{SAT Writing Craft and Structure (CS).}  
Evaluates understanding of vocabulary in context and text organization. Students must select appropriate words based on context, analyze how ideas develop across paragraphs, and recognize how structure enhances effectiveness. Requires understanding both denotative and connotative meanings while evaluating how word choice affects tone, style, and precision of expression.

\begin{tcolorbox}[%
  colback=white,
  colframe=black,
  width=\columnwidth,
  boxrule=0.5pt,
  arc=3pt,
  left=5pt,
  right=5pt,
  boxsep=0pt,
  top=3pt,
  bottom=3pt
]
Sentence: The scientist's findings were \_\_\_\_\_, shedding light on the mysterious behaviors of subatomic particles that had puzzled physicists for decades\dots\\
A. groundbreaking\\
B. world-class\\
C. groundbreaking and thrilling\\
D. groundbreaking, thrilling
\end{tcolorbox}

\noindent\textbf{SAT Expression of Ideas (EI).}  
Focuses on writing effectiveness and logical flow. Students evaluate and improve coherence, cohesion, and clarity by combining sentences, reorganizing information, or modifying details. Requires determining relevance, identifying optimal placement for new information, and understanding how structural changes affect meaning and emphasis. Tests ability to develop logically connected ideas with appropriate support.

\begin{tcolorbox}[
    colback=white,
    colframe=black,
    width=\columnwidth,
    boxrule=0.5pt,
    arc=3pt,
    left=5pt,
    right=5pt,
    boxsep=0pt,
    top=3pt,
    bottom=3pt,
]
Revision Task: Improve the coherence of the following sentence pair to create a more logical flow:

“I love classical music. Beethoven’s symphonies are my favorite.”

Possible revisions:

A. I love classical music, especially Beethoven’s symphonies, which are my favorite.

B. Because I love classical music, Beethoven’s symphonies are my favorite.

C. I love classical music; indeed, Beethoven’s symphonies are my favorite.

D. No change necessary.
\end{tcolorbox}

\noindent\textbf{SAT English Conventions (EC).}  
Assesses command of standard English grammar, punctuation, and sentence structure. Topics include verb tense/agreement, pronoun usage, parallel structure, modifier placement, and appropriate punctuation. Students must identify and correct errors in sentences or paragraphs. Evaluates practical application of grammatical rules rather than theoretical knowledge.

\begin{tcolorbox}[
    colback=white,
    colframe=black,
    width=\columnwidth,
    boxrule=0.5pt,
    arc=3pt,
    left=5pt,
    right=5pt,
    boxsep=0pt,
    top=3pt,
    bottom=3pt,
]

Sentence: Each of the students (have / has) submitted their essay on time, and the teacher (is / are) pleased with the quality of work...

A. have, is \\
B. has, is \\
C. have, are \\
D. has, are 
\end{tcolorbox}

\noindent\textbf{SAT Math Algebra (AG).}  
Tests ability to work with algebraic expressions, equations, inequalities, and functions. Requires solving linear/quadratic equations, manipulating expressions, understanding variable relationships, and analyzing functions. Students must apply algebraic concepts to model real-world situations and connect symbolic representations with graphs. Assesses both procedural fluency and conceptual understanding.
\begin{tcolorbox}[
    colback=white,
    colframe=black,
    width=\columnwidth,
    boxrule=0.5pt,
    arc=3pt,
    left=5pt,
    right=5pt,
    boxsep=0pt,
    top=3pt,
    bottom=3pt,
]
Solve for \(x\): \(\frac{2x-3}{x+1} = 4\)

A. x = 7 \\
B. x = -7 \\
C. x = 7/3 \\
D. x = -7/3 \\
E. No solution exists
\end{tcolorbox}
\noindent\textbf{SAT Data Analysis (DA).}  
Evaluates interpretation of various data forms. Students analyze ratios, rates, percentages, proportions, and probabilities while interpreting information from tables, charts, and graphs. Requires understanding statistical concepts (mean, median, mode) and using data to draw conclusions. Tests quantitative literacy skills needed for interpreting real-world numerical information.

\begin{tcolorbox}[
    colback=white,
    colframe=black,
    width=\columnwidth,
    boxrule=0.5pt,
    arc=3pt,
    left=5pt,
    right=5pt,
    boxsep=0pt,
    top=3pt,
    bottom=3pt,
]
A circle graph shows that 30\% of students prefer tea, 50\% coffee, and 20\% water. If 200 students were surveyed, how many prefer coffee?  

A. 60\\
B. 100\\
C. 40\\
D. 50\\
E. Cannot be determined from the information given
\end{tcolorbox}
 
\noindent\textbf{SAT Geometry \& Trigonometry (GT).}  
Covers geometric figures, coordinate geometry, and trigonometric relationships. Questions involve angles, lines, polygons, circles, 3D figures, coordinate systems, and trigonometric functions. Students calculate areas, perimeters, volumes, and distances while applying properties of various shapes. Tests spatial reasoning and connection of algebraic and geometric representations.

\begin{tcolorbox}[
    colback=white,
    colframe=black,
    width=\columnwidth,
    boxrule=0.5pt,
    arc=3pt,
    left=5pt,
    right=5pt,
    boxsep=0pt,
    top=3pt,
    bottom=3pt,
]
In right triangle \(ABC\) with right angle at \(C\), if \(AC=3\) and \(BC=4\), what is \(\sin A\)?  

A. 3/5 \\
B. 4/5 \\
C. 3/4\\
D. 4/3\\
E. 5/3
\end{tcolorbox}

\noindent\textbf{GRE Text Completion (TC).}  
Assesses vocabulary and comprehension by requiring completion of blanks in short passages. Students must understand author's intent, logical relationships between sentences, and overall context. Difficulty increases with multiple blanks where choices must work cohesively. Tests vocabulary breadth and understanding of how words function within complex contexts.

\begin{tcolorbox}[%
  colback=white,
  colframe=black,
  width=\columnwidth,
  boxrule=0.5pt,
  arc=3pt,
  left=5pt,
  right=5pt,
  boxsep=0pt,
  top=3pt,
  bottom=3pt
]
Though praised for its \_\_\_\_\_\ innovations, the clock's design was too \_\_\_\_\_ to gain widespread adoption among consumers who valued simplicity and ease of use...

A. aesthetic ... cumbersome \\
B. mechanical ... simplified \\
C. technological ... intuitive \\
D. functional ... intricate \\
E. rudimentary ... complex
\end{tcolorbox}

\noindent\textbf{GRE Sentence Equivalence (SE).}  
Requires selecting two words that create sentences with equivalent meanings when inserted. Students must identify words that produce the same overall meaning in context, understanding subtle connotative differences. Tests vocabulary depth, contextual word usage, and ability to maintain consistent meaning across different word choices.
\begin{tcolorbox}[%
    colback=white,
    colframe=black,
    width=\columnwidth,
    boxrule=0.5pt,
    arc=3pt,
    left=5pt,
    right=5pt,
    boxsep=0pt,
    top=3pt,
    bottom=3pt,
]
Her lecture was so \_\_\_\_\_ that many students struggled to stay awake.\\  
A. engaging  \\
B. soporific  \\
C. bewildering  \\
D. tedious  \\
E. stimulating \\
F. monotonous
\end{tcolorbox}

\noindent\textbf{GRE Reading Comprehension (RC).}  
Tests analysis and interpretation of complex academic passages. Students identify main ideas, recognize explicit statements, make inferences, understand author's purpose, and evaluate arguments. Requires handling sophisticated vocabulary and complex sentence structures while synthesizing information across passages and drawing conclusions from implied content.

\begin{tcolorbox}[
    colback=white,
    colframe=black,
    width=\columnwidth,
    boxrule=0.5pt,
    arc=3pt,
    left=5pt,
    right=5pt,
    boxsep=0pt,
    top=3pt,
    bottom=3pt,
]
Passage: Advances in CRISPR technology have opened new avenues in gene therapy, offering unprecedented precision in modifying DNA sequences. Unlike earlier gene-editing methods that often resulted in unintended modifications, CRISPR-Cas9 allows scientists to target specific sections of genetic code with remarkable accuracy...

Question: The passage suggests that CRISPR's main advantage over previous gene-editing methods is its ability to:\\
A. Work faster than other methods \\
B. Target specific sections of genetic code with high accuracy\\
C. Completely eliminate the risk of unintended modifications\\
D. Address a wider range of medical conditions\\
E. Bypass ethical concerns associated with genetic manipulation
\end{tcolorbox}

\noindent\textbf{GRE Quantitative Comparison (QC).}  
Presents two quantities for comparison of relative size. Tests conceptual understanding over computational ability as students analyze information, identify mathematical relationships, and determine if enough information exists to establish definitive relationships. Requires creative approaches, estimation skills, and recognition of information adequacy without necessarily performing complex calculations.

\begin{tcolorbox}[
    colback=white,
    colframe=black,
    width=\columnwidth,
    boxrule=0.5pt,
    arc=3pt,
    left=5pt,
    right=5pt,
    boxsep=0pt,
    top=3pt,
    bottom=3pt,
]
Quantity A: \(2^{10}\)  
\hspace{20pt} Quantity B: \(10^3\)  

A. Quantity A is greater \\
B. Quantity B is greater\\
C. The two quantities are equal\\
D. The relationship cannot be determined from the information given
\end{tcolorbox}
\noindent\textbf{GRE Numeric Entry (NE).}  
 Requires calculating exact answers without multiple-choice options. Tests ability to perform calculations accurately and follow procedures correctly without answer verification. Assesses computational skills, problem-solving strategies, and work with various numerical forms. Demands confidence in mathematical procedures and attention to units and precision.

\begin{tcolorbox}[
    colback=white,
    colframe=black,
    width=\columnwidth,
    boxrule=0.5pt,
    arc=3pt,
    left=5pt,
    right=5pt,
    boxsep=0pt,
    top=3pt,
    bottom=3pt,
]
If a tank is filled at a constant rate and holds 65 gallons in 10 minutes, how many gallons per minute are being added to the tank? If the answer is a fraction, enter as a decimal.

Answer box: \_\_\_\_\_
\end{tcolorbox}

\noindent\textbf{GRE Data Interpretation (DI).}  
Assesses ability to analyze and interpret data in graphs, tables, or charts. Students extract information, perform calculations, recognize patterns, and draw conclusions. Requires comparing data points, calculating percentages or rates of change, and making predictions. Tests quantitative literacy and ability to work with real-world data representations.

\begin{tcolorbox}[
    colback=white,
    colframe=black,
    width=\columnwidth,
    boxrule=0.5pt,
    arc=3pt,
    left=5pt,
    right=5pt,
    boxsep=0pt,
    top=3pt,
    bottom=3pt,
]
Table: Quarterly profits (in \$M) for Company X: \\
Q1: 10 \\
Q2: 15 \\
Q3: 12 \\
Q4: 18 

Question: In which quarter did Company X see the greatest increase in profit over the previous quarter?
A. Q1 \\
B. Q2 \\
C. Q3 \\
D. Q4 \\
E. Cannot be determined from the information given
\end{tcolorbox}

\noindent\textbf{GMAT Critical Reasoning (CR).}  
Evaluates analysis of argument structure, validity, and logical coherence. Students identify premises, conclusions, and assumptions while distinguishing relevant information and recognizing logical flaws. Often uses business scenarios requiring understanding of causation vs. correlation and sample representativeness. Tests analytical thinking crucial for business decision-making.

\begin{tcolorbox}[
    colback=white,
    colframe=black,
    width=\columnwidth,
    boxrule=0.5pt,
    arc=3pt,
    left=5pt,
    right=5pt,
    boxsep=0pt,
    top=3pt,
    bottom=3pt,
]
Argument: Because sales rose by 15\% last quarter immediately following the implementation of our new marketing strategy, the new marketing strategy must be effective and should be continued without modifications in the upcoming fiscal year.

Which of the following, if true, most weakens this conclusion? \\

A. The company's main competitor went out of business during the same quarter. \\
B. The company introduced a popular new product line at the beginning of the quarter. \\
C. Other companies using similar marketing strategies saw comparable increases in sales. \\
D. The marketing strategy cost more to implement than initially projected. \\
E. Industry sales overall rose by 20\% during the same period due to seasonal factors.
\end{tcolorbox}

\noindent\textbf{GMAT Reading Comprehension (RC).}  
Similar to GRE but with more business focus. Tests understanding of complex written material, identification of main ideas, inference-making, and logical structure recognition. Passages often discuss business strategies or economic concepts. Assesses ability to distinguish stated from implied information and evaluate argument strength.

\begin{tcolorbox}[
    colback=white,
    colframe=black,
    width=\columnwidth,
    boxrule=0.5pt,
    arc=3pt,
    left=5pt,
    right=5pt,
    boxsep=0pt,
    top=3pt,
    bottom=3pt,
]
Passage: Global coffee consumption has doubled in the past decade, driven primarily by emerging markets in Asia where a growing middle class has embraced Western consumption patterns. China, traditionally a tea-drinking nation, has seen coffee consumption grow at 15\% annually, compared to global growth of 2.5\%...

Question: The author primarily discusses which factor driving coffee demand? \\
A. Changes in consumer taste preferences \\
B. Economic development and social status in emerging markets \\
C. Declining popularity of traditional tea consumption \\
D. Marketing strategies of Western coffee companies \\
E. Health benefits associated with coffee consumption
\end{tcolorbox}

\noindent\textbf{GMAT Problem Solving (PS).}  
Tests mathematical knowledge across arithmetic, algebra, geometry, and statistics. Students determine problem requirements, identify relevant information, select appropriate techniques, and calculate accurately. Requires translating word problems into mathematical expressions and interpreting solutions in context, often in business-related scenarios.

\begin{tcolorbox}[
    colback=white,
    colframe=black,
    width=\columnwidth,
    boxrule=0.5pt,
    arc=3pt,
    left=5pt,
    right=5pt,
    boxsep=0pt,
    top=3pt,
    bottom=3pt,
]
If \(x+y=10\) and \(xy=21\), what is \(x^2 + y^2\)?  \\
A. 52 \\
B. 58 \\
C. 100 \\
D. 121 \\
E. 142
\end{tcolorbox}

\noindent\textbf{GMAT Data Sufficiency (DS).}  
A unique format assessing analytical thinking over computation. Students determine whether statements provide sufficient information to answer questions without actually solving problems. Requires evaluating statement sufficiency individually and collectively while understanding necessary vs. sufficient conditions and recognizing implied information.

\begin{tcolorbox}[
    colback=white,
    colframe=black,
    width=\columnwidth,
    boxrule=0.5pt,
    arc=3pt,
    left=5pt,
    right=5pt,
    boxsep=0pt,
    top=3pt,
    bottom=3pt,
]
Question: Is \(x>5\)?  
(1) \(2x>10\)  
(2) \(x^2>25\)  

A. Statement (1) ALONE is sufficient, but statement (2) ALONE is not sufficient. \\
B. Statement (2) ALONE is sufficient, but statement (1) ALONE is not sufficient.\\
C. BOTH statements TOGETHER are sufficient, but NEITHER statement ALONE is sufficient.\\
D. EACH statement ALONE is sufficient.\\
E. Statements (1) and (2) TOGETHER are NOT sufficient.
\end{tcolorbox}

\noindent\textbf{GMAT Integrated Reasoning (IR).}  
Tests analysis of information from multiple sources and formats. Students interpret tables, graphs, and text while evaluating multiple information sources and solving multi-step problems. Includes multi-source reasoning, graphics interpretation, two-part analysis, and table analysis. Assesses skills for data-driven business decisions.

\begin{tcolorbox}[
    colback=white,
    colframe=black,
    width=\columnwidth,
    boxrule=0.5pt,
    arc=3pt,
    left=5pt,
    right=5pt,
    boxsep=0pt,
    top=3pt,
    bottom=3pt,
]
Table: 
Region A sales (in millions): Year 1: \$120, Year 2: \$132, Year 3: \$145
Region B sales (in millions): Year 1: \$90, Year 2: \$101, Year 3: \$114

Question: Which region's compound annual growth rate exceeded 5\% over the three-year period?\\
A. Region A only\\
B. Region B only\\
C. Both Region A and Region B\\
D. Neither Region A nor Region B
\end{tcolorbox}
\noindent\textbf{TOEFL Reading Factual Information (FI).}  
Evaluates ability to identify explicitly stated facts in academic texts. Students locate specific information, distinguish it from similar content, and understand its contextual significance. Requires processing academic vocabulary and syntax while focusing on directly stated rather than implied information.

\begin{tcolorbox}[
    colback=white,
    colframe=black,
    width=\columnwidth,
    boxrule=0.5pt,
    arc=3pt,
    left=5pt,
    right=5pt,
    boxsep=0pt,
    top=3pt,
    bottom=3pt,
]
Passage: Canada's boreal forest covers nearly one-third of its land area, spanning from Yukon to Newfoundland and Labrador. This vast ecosystem, dominated by coniferous trees, contains more than 1.5 million lakes and is home to endangered species such as the woodland caribou...

Question: According to the passage, what fraction of Canada is covered by the boreal forest? \\
A. One-quarter \\
B. One-third \\
C. One-half \\
D. Two-thirds
\end{tcolorbox}

\noindent\textbf{TOEFL Inference \& Reference (IR).}  
Tests understanding of implied information and referential relationships. Students draw logical conclusions from provided information, understand unstated relationships between ideas, and track references through pronouns and demonstratives. Assesses deeper comprehension including reading between lines and connecting ideas across text sections.

\begin{tcolorbox}[
    colback=white,
    colframe=black,
    width=\columnwidth,
    boxrule=0.5pt,
    arc=3pt,
    left=5pt,
    right=5pt,
    boxsep=0pt,
    top=3pt,
    bottom=3pt,
]
Passage: "The experiment failed again, prompting the research team to reconsider their methodology. Dr. Chen suggested they should explore alternative approaches that had shown promise in similar contexts."

Question: What does "again" imply about previous attempts?\\
A. This was the first time the experiment had been conducted. \\
B. Previous attempts had been successful.\\
C. Previous attempts had also failed.\\
D. The team had never tried this experiment before.
\end{tcolorbox}

\noindent\textbf{TOEFL Text \& Sentence (TS).}  
Evaluates various aspects of textual understanding including summarizing and sentence relationships. Tasks include inserting sentences appropriately, creating cohesive summaries, simplifying complex sentences, or identifying sentence functions. Tests advanced language processing including understanding textual connections, organizational structure, and purpose of different elements.
\begin{tcolorbox}[
    colback=white,
    colframe=black,
    width=\columnwidth,
    boxrule=0.5pt,
    arc=3pt,
    left=5pt,
    right=5pt,
    boxsep=0pt,
    top=3pt,
    bottom=3pt,
]
Original: Because of the severe weather conditions, we decided to cancel the outdoor concert scheduled for tomorrow evening.

Task: Combine into one sentence without changing meaning, beginning with "The outdoor concert..."
A. The outdoor concert scheduled for tomorrow evening we decided to cancel because of the severe weather conditions. \\
B. The outdoor concert scheduled for tomorrow evening was decided to be  cancelled by us because of the severe weather conditions. \\
C. The outdoor concert scheduled for tomorrow evening has been cancelled due to the severe weather conditions.\\
D. The outdoor concert, because of the severe weather conditions, scheduled for tomorrow evening we decided to cancel.
\end{tcolorbox}
\noindent\textbf{TOEFL Listening Factual Information (FI).}  
Assesses comprehension of spoken academic content. Students identify explicitly stated information, distinguish between similar details, and recognize contextual significance. Requires processing natural-speed academic English despite accent variations while maintaining focus during extended listening passages.

\begin{tcolorbox}[
    colback=white,
    colframe=black,
    width=\columnwidth,
    boxrule=0.5pt,
    arc=3pt,
    left=5pt,
    right=5pt,
    boxsep=0pt,
    top=3pt,
    bottom=3pt,
]
Transcript: "Good morning, class. Today's lecture will cover photosynthesis, the process by which plants convert light energy into chemical energy. We'll first discuss the light-dependent reactions that occur in the thylakoid membrane, followed by the Calvin cycle that takes place in the stroma..."

Question: What topic will the lecture cover? \\
A. Cell respiration \\
B. Plant reproduction \\
C. Photosynthesis \\
D. Genetic engineering
\end{tcolorbox}

\noindent\textbf{TOEFL Listening Inference (IF).}  
Tests understanding of implied meanings in spoken content. Students interpret speaker's tone, infer unstated opinions, understand implied connections, and determine purpose of specific statements. Requires comprehending not just words but also intonation and emphasis. Assesses ability to understand nuanced academic communication.
\begin{tcolorbox}[
    colback=white,
    colframe=black,
    width=\columnwidth,
    boxrule=0.5pt,
    arc=3pt,
    left=5pt,
    right=5pt,
    boxsep=0pt,
    top=3pt,
    bottom=3pt,
]
Speaker: "I suppose we could try that method, if all our other options have been exhausted. It's not my first choice, but at this point, we might not have many alternatives left."

Question: What does the speaker's tone suggest about their enthusiasm for the proposed method?
A. They are excited to try something new \\
B. They are reluctant but resigned to trying it \\
C. They believe it is the best available option \\
D. They are confident it will succeed
\end{tcolorbox}
\noindent\textbf{IELTS Reading Identifying Information (II).}  
Evaluates whether statements match textual information. Students determine if statements are True (matching), False (contradicting), or Not Given (not addressed). Requires careful reading to distinguish between explicit, inferable, and absent information without introducing outside knowledge.
\begin{tcolorbox}[
    colback=white,
    colframe=black,
    width=\columnwidth,
    boxrule=0.5pt,
    arc=3pt,
    left=5pt,
    right=5pt,
    boxsep=0pt,
    top=3pt,
    bottom=3pt,
]
Passage: "Many cities have embraced rooftop gardens as a sustainable solution to multiple urban challenges. These green spaces not only provide fresh produce for local communities but also help mitigate the urban heat island effect by absorbing sunlight that would otherwise be converted to heat..."

Statement: "The author believes urban gardens are ineffective at addressing environmental challenges."

Is the statement True, False, or Not Given?
\end{tcolorbox}

\noindent\textbf{IELTS Matching Sentence (MS).}  
Tests ability to connect related information pieces. Students match headings with paragraphs, sentence beginnings with endings, or statements with speakers. Requires understanding paragraph main ideas, sentence logic, and information relationships while processing content across multiple text sections.

\begin{tcolorbox}[
    colback=white,
    colframe=black,
    width=\columnwidth,
    boxrule=0.5pt,
    arc=3pt,
    left=5pt,
    right=5pt,
    boxsep=0pt,
    top=3pt,
    bottom=3pt,
]
Complete the following sentence with the most appropriate ending from the list below:

"Fossil fuels are being replaced by renewable sources..." 

A. ...because they are more sustainable and environmentally friendly. \\
B. ...despite their continued dominance in global energy markets. \\
C. ...although the transition is happening more slowly than many scientists recommend.\\
D. ...particularly in developing economies seeking to reduce energy costs.\\
E. ...which has caused significant economic disruption in traditional energy sectors.
\end{tcolorbox}

\noindent\textbf{IELTS Completion (CP).}  
Assesses ability to locate and transfer specific information to complete sentences, summaries, or diagrams. Students identify relevant details and transfer them accurately, often verbatim. Requires understanding text structure for efficient information location while recognizing synonyms and paraphrased content.

\begin{tcolorbox}[
    colback=white,
    colframe=black,
    width=\columnwidth,
    boxrule=0.5pt,
    arc=3pt,
    left=5pt,
    right=5pt,
    boxsep=0pt,
    top=3pt,
    bottom=3pt,
]
Summary: "The Sahara is the world's \_\_\_\_\_ desert, covering approximately \_\_\_\_\_ million square kilometers across North Africa, from the Atlantic Ocean to the Red Sea. Its name comes from the Arabic word meaning \_\_\_\_\_."
Words to choose from:\\
A. largest, 9.2, "desert"\\
B. hottest, 8.7, "sand"\\
C. oldest, 7.5, "wilderness"\\
D. driest, 6.3, "emptiness"
\end{tcolorbox}

\noindent\textbf{IELTS Listening Identification \& Matching (IM).}  
Tests identification of specific spoken information and category matching. Students listen for details like names, numbers, and facts then select correct options. Requires processing natural-speed English despite distractions or accent variations while distinguishing between similar-sounding choices.

\begin{tcolorbox}[
    colback=white,
    colframe=black,
    width=\columnwidth,
    boxrule=0.5pt,
    arc=3pt,
    left=5pt,
    right=5pt,
    boxsep=0pt,
    top=3pt,
    bottom=3pt,
]
Audio transcript: "Welcome to our university orientation. The main campus tour will begin at the Student Center at quarter past nine. Please arrive at least ten minutes early to collect your information packets..."

Question: What time does the campus tour start? \\
A. 9:00 \\
B. 9:15 \\
C. 9:30 \\
D. 10:15
\end{tcolorbox}

\noindent\textbf{IELTS Completion \& Labeling (CL).}  
Evaluates ability to listen for specific information to complete sentences, notes, or diagrams. Students identify and record specific details, often verbatim. Requires focused listening, accurate information processing, and simultaneous writing. Tests note-taking skills needed for educational and professional contexts.
\begin{tcolorbox}[
    colback=white,
    colframe=black,
    width=\columnwidth,
    boxrule=0.5pt,
    arc=3pt,
    left=5pt,
    right=5pt,
    boxsep=0pt,
    top=3pt,
    bottom=3pt,
]
[Audio describes the parts of a flower and their functions]

Diagram: Label the parts of a flower shown in the image using words from the recording:\\

1. \_\_\_\_ (outer protective layer)\\
2. \_\_\_\_ (colorful structures that attract pollinators)\\
3. \_\_\_\_ (male reproductive part containing pollen)\\
4. \_\_\_\_ (female reproductive structure)\\
5. \_\_\_\_ (produces seeds when fertilized)
\end{tcolorbox}

\noindent\textbf{IELTS Short Answer (SA).}  
Tests listening for specific information and providing concise answers using the recording's words. Students identify relevant details and express them within word limits. Requires understanding question focus, quick information processing, and appropriate word selection. Assesses both receptive and productive language skills.

\begin{tcolorbox}[
    colback=white,
    colframe=black,
    width=\columnwidth,
    boxrule=0.5pt,
    arc=3pt,
    left=5pt,
    right=5pt,
    boxsep=0pt,
    top=3pt,
    bottom=3pt,
]
Audio transcript: "For our upcoming science class field trip next Thursday, we'll be visiting the botanical gardens on the north side of the city. Please remember to bring your permission slips, a notebook, appropriate footwear, and a packed lunch..."

Questions: \\
1. Where is the field trip? (Answer in no more than THREE words) \\
2. What day will the field trip take place? (Answer in no more than TWO words) \\
3. What time will students return to school? (Answer in no more than TWO words)
\end{tcolorbox}

\section{Pedagogical Alignment of the Cognitive Trajectory with Human Test-Taking Strategies}
\label{app:breakdown}

To ensure that \bench serves as a valid diagnostic tool, we must justify that our formalized cognitive trajectory mirrors the actual cognitive strategies adopted by high-performing human test-takers. Our design is grounded in well-documented findings from standardized test preparation curricula, Cognitive Diagnosis Models (CDMs), and empirical think-aloud studies of student behaviors during exam practice \citep{loken2004online,collegeboard2025_cognitive,johnstone2006thinkaloud}. Below, we elaborate on how human heuristics map to our three pedagogical  categories (defined in \S\ref{ssec:task}).

\subsection*{Category I: Knowledge-Intensive Retrieval (Lexical \& Structural)}
\noindent {\bf Cognitive Trajectories:} $c_{\text{syntax\_parse}} \rightarrow c_{\text{rule\_match}} \rightarrow c_{\text{predict}}$ \quad | \quad $c_{\text{semantic\_parse}} \rightarrow c_{\text{logical\_scope}} \rightarrow c_{\text{resolve}}$ \
In grammar and vocabulary-oriented tasks (e.g., SAT Writing, GRE Text Completion), human test-takers do not evaluate sentences holistically; instead, they engage in mental `chunking.'' They first parse the syntactic or semantic structure ($c_{\text{syntax\_parse}}$), identifying structural anchors like subject-verb pairs or contrast markers (e.g., `although''). Next, they localize the logical scope by matching the text against internalized grammatical rules or expected vocabulary contexts ($c_{\text{rule\_match}}$). Finally, they predict the missing element or resolve the meaning ($c_{\text{predict}}$) before looking at the multiple-choice options. Our cognitive trajectory explicitly encodes this systematic progression, penalizing models that attempt to guess the final word without first demonstrating structural awareness.

\subsection*{Category II: Reasoning-Intensive Execution (Multimodal \& Quantitative)}
\noindent {\bf Cognitive Trajectories:} $c_{\text{analytical\_goal}} \rightarrow c_{\text{visual\_parse}} \rightarrow c_{\text{analyze}}$ \quad | \quad $c_{\text{model}} \rightarrow c_{\text{formulate\_eq}} \rightarrow c_{\text{compute}}$ \
Quantitative and multimodal questions (e.g., SAT Math, GRE Data Interpretation) require strict procedural execution. Empirical studies show that human test-takers who rush directly into computation are highly prone to errors. Effective problem-solvers first establish an analytical goal ($c_{\text{analytical\_goal}}$) or translate a natural language word problem into a conceptual model ($c_{\text{model}}$). For multimodal tasks, test-prep curricula heavily emphasize reading the axes, legends, and units of a chart ($c_{\text{visual\_parse}}$) prior to analyzing the data. For mathematical tasks, humans translate their conceptual model into a rigorous algebraic representation ($c_{\text{formulate\_eq}}$) before engaging in symbolic manipulation or arithmetic execution ($c_{\text{compute}}$). Our cognitive framework enforces this exact procedural separation, allowing us to diagnose whether an LLM fails due to a misunderstanding of the mathematical text ($c_{\text{model}}$) or a pure arithmetic hallucination ($c_{\text{compute}}$).

\subsection*{Category III: Hybrid Integration (Semantic Extraction \& Inference)}
\noindent {\bf Cognitive Trajectories:} $c_{\text{intent}} \rightarrow c_{\text{perceive\_text}} \rightarrow c_{\text{extract}}$ \quad | \quad $c_{\text{entities}} \rightarrow c_{\text{constraints}} \rightarrow c_{\text{evaluate}}$ \
For tasks requiring dense reading comprehension or logical deduction (e.g., TOEFL Reading, GMAT Critical Reasoning), human test-takers employ targeted scanning and hypothesis testing. They begin by identifying the core subject or intent of the question stem ($c_{\text{intent}}$) to avoid wasting working memory on irrelevant text. They then scan the passage or audio mental representation for relevant discourse ($c_{\text{perceive\_text}}$). In comparative logic tasks, human strategies explicitly emphasize identifying the entities to be compared ($c_{\text{entities}}$) and strictly mapping given constraints ($c_{\text{constraints}}$) before evaluating sufficiency ($c_{\text{evaluate}}$). This mirrors heuristic strategies taught in test prep, where students are trained to systematically test conditions rather than rely on intuition. By isolating these nodes, our cognitive framework evaluates whether an LLM can mimic this disciplined information retrieval and constraint satisfaction.

\subsection*{Summary of Pedagogical Fidelity}
Across all categories, the cognitive trajectory decomposes EST problem-solving into sequential operations that are (1) cognitively plausible, (2) perfectly aligned with established test-preparation strategies, and (3) empirically observable in human behavior. Consequently, LLMs evaluated using our Stepwise Execution Score (SES) are not merely graded on answer accuracy, but on their ability to exhibit an explainable, human-aligned reasoning process suitable for real-world educational deployment.

\section{Prompt Structure Used in Evaluation}
\label{app:prompt}

This appendix summarizes the skeleton templates for three prompt-engineering paradigms used in our educational reasoning experiments. Each template is presented with an explanation followed by a concrete example in a highlighted box.

\subsection*{1. In-Context Learning (ICL) Prompt Structure} 

Provides the model with solved examples to prime analogous problem solving:
\begin{itemize}
    \item Multiple exemplars demonstrating the problem–solution pattern
    \item Graduated difficulty progression across examples
    \item Explicit identification of transferable patterns in each exemplar
    \item Strategic selection of examples to highlight different facets of the problem type
    \item Clear demarcation between example set and the target query
\end{itemize}

\begin{tcolorbox}[
    colback=white, 
    colframe=black, 
    width=\columnwidth, 
    boxrule=0.5pt, 
    arc=3pt, 
    left=5pt, 
    right=5pt, 
    boxsep=0pt, 
    top=3pt, 
    bottom=3pt,
]
\textbf{Example:} For a simple SAT algebra problem—"Solve for $x$: $(x-2)(x+3)=0$":
\begin{tabular}{@{}p{0.02\columnwidth}@{}p{0.92\columnwidth}@{}}
    1. & Input: "Solve $(x-2)(x+3)=0$" \\
       & Output: "Step 1: Recognize factors imply solutions x=2 or x=-3.Step 2: List solutions: 2, -3."
\end{tabular}
\end{tcolorbox}

\subsection*{2. Chain-of-Thought (CoT) Prompt Structure} 

Guides the model through a step-by-step reasoning process:
\begin{itemize}
    \item Instruction to decompose the task into ordered steps
    \item Explicit requests for intermediate calculations or justifications
    \item Structured step-labeling conventions (e.g., ``Step 1: \ldots'', ``Step 2: \ldots'')
    \item Prompts for linking each step's result to the next
    \item Final summary or explicit statement of the conclusion
\end{itemize}

\begin{tcolorbox}[
    colback=white, 
    colframe=black, 
    width=\columnwidth, 
    boxrule=0.5pt, 
    arc=3pt, 
    left=5pt, 
    right=5pt, 
    boxsep=0pt, 
    top=3pt, 
    bottom=3pt,
]
\textbf{Example:} Compute the area of a triangle with base 5 and height 8:
\begin{enumerate}
    \item Step 1: Write formula $A=\frac{1}{2} \times \text{base} \times \text{height}$. \\
    \item Step 2: Substitute values: $A=\frac{1}{2} \times 5 \times 8$. \\
    \item Step 3: Calculate: $A=20$. \\
    \item Conclusion: The area is 20.
\end{enumerate}
\end{tcolorbox}

\subsection*{3. Tree-of-Thought (ToT) Prompt Structure} 

Encourages exploration of multiple reasoning branches before selecting the optimal path:
\begin{itemize}
    \item Generate a set of candidate ``thoughts'' for the first reasoning step
    \item For each candidate, expand into next-level thoughts, optionally scoring or pruning
    \item Continue branching until a termination criterion is met (depth limit or score threshold)
    \item Compare complete reasoning chains and select the highest-scoring sequence
    \item Output the final answer along with the chosen reasoning path
\end{itemize}

\begin{tcolorbox}[
    colback=white, 
    colframe=black, 
    width=\columnwidth, 
    boxrule=0.5pt, 
    arc=3pt, 
    left=5pt, 
    right=5pt, 
    boxsep=0pt, 
    top=3pt, 
    bottom=3pt,
]
\textbf{Example:} For solving $3x^2 - 10x + 7 = 0$, explore:
\begin{itemize}
    \item Thought A: Factorization approach
    \item Thought B: Quadratic formula
    \item Thought C: Vieta's formulas
\end{itemize}
Evaluate efficiency and choose Vieta's: sum of roots $= \frac{10}{3}$, product of roots $= \frac{7}{3}$.
\end{tcolorbox}

\subsection{Experimental Setting}
\label{app:setting}

This section lists the experimental settings used in this study.

\begin{table*}[h]
\centering
\small
\caption{LLM query hyperparameters used during all experiments.}
\begin{tabular}{lcc}
\toprule
\textbf{Hyperparameter} & \textbf{Value} & \textbf{Description} \\
\midrule
Temperature     & 0.7      & Controls randomness in generation \\
Top-$p$ (nucleus sampling) & 0.95     & Probability mass for sampling \\
Max tokens      & 2048     & Maximum number of tokens to generate \\
Stop sequences  & [\texttt{"\textbackslash n"}, \texttt{"Q:"}] & Used to truncate responses \\
Prompt format   & CoT, CoT-SC, ToT & Prompting strategy used in Section \ref{sec:expt} \\
\bottomrule
\end{tabular}
\label{tab:hyperparams}
\end{table*}

{\bf Computational Resources.} All experiments were conducted on a high-performance computing server equipped with six NVIDIA RTX 6000 Ada Generation GPUs, each with 49 GB of dedicated VRAM. The system utilized CUDA version 12.8 and NVIDIA driver version 570.124.06. These GPUs supported parallel execution of model querying, evaluation, and tool-augmented tasks across our benchmark datasets. The hardware configuration ensured sufficient memory bandwidth and processing capability to accommodate large-scale inference, particularly for multimodal tasks and multi-sample prompting strategies such as CoT-SC and ToT. No resource-related constraints were encountered during experimentation.

\section{Additional Experimental Results and Discussions for RQ$_1$}
\label{app:expt-more-rq1}

This section presents additional details and experimental results that complement the main evaluation in \S\ref{ssec:expt-main}. These supplementary findings, together with what has been presented in previous sections, offer comprehensive insights into LLMs capabilities across different EST tasks.

\subsection{Complementing experimental results}
\label{app:add-expt-rq1}

\begin{table*}[!t]
\caption{Standard deviations of performance across five runs. Human testers generally show higher variability, though LLMs also fluctuate, especially on multimodal and quantitative tasks.}
\resizebox{\textwidth}{!}{
\centering
\begin{tabular}{l|c|ccc|ccc|ccc|ccc|ccc|ccc}
\toprule
\multirow{2.5}{*}{\bf Task} & \multirow{2.5}{*}{\bf Human} 
& \multicolumn{3}{c}{\bf GPT-4V} 
& \multicolumn{3}{c}{\bf GPT-5} 
& \multicolumn{3}{c}{\bf Claude-Sonnet-4} 
& \multicolumn{3}{c}{\bf Llama-4-Scout-17B} 
& \multicolumn{3}{c}{\bf Qwen-VL-Max} 
& \multicolumn{3}{c}{\bf Gemini-2.5} \\
\cmidrule(lr){3-5}\cmidrule(lr){6-8}\cmidrule(lr){9-11}\cmidrule(lr){12-14}\cmidrule(lr){15-17}\cmidrule(lr){18-20}
& & ICL & CoT & ToT & ICL & CoT & ToT & ICL & CoT & ToT & ICL & CoT & ToT & ICL & CoT & ToT & ICL & CoT & ToT \\
\midrule
\multicolumn{20}{c}{\it SAT} \\
\midrule
II & 2.8 & 0.3 & 0.6 & 1.1 & 0.4 & 1.2 & 1.5 & 0.7 & 0.9 & 1.8 & 2.9 & 3.6 & 4.8 & 1.5 & 1.9 & 2.2 & 0.8 & 2.0 & 2.6 \\
CS & 5.2 & 0.5 & 1.1 & 1.8 & 0.7 & 1.0 & 1.5 & 2.2 & 3.0 & 4.1 & 3.6 & 4.5 & 5.1 & 1.2 & 1.6 & 2.2 & 0.9 & 1.5 & 2.1 \\
EI & 8.5 & 0.4 & 0.9 & 1.3 & 0.5 & 0.8 & 1.2 & 1.7 & 2.5 & 3.8 & 2.8 & 3.6 & 4.4 & 1.1 & 1.5 & 1.9 & 0.6 & 1.4 & 1.8 \\
EC & 4.9 & 0.3 & 0.7 & 1.0 & 0.4 & 0.9 & 1.4 & 1.6 & 2.3 & 3.5 & 3.2 & 3.8 & 4.7 & 0.9 & 1.2 & 1.6 & 0.8 & 1.3 & 2.0 \\
AG & 14.2 & 1.2 & 2.1 & 2.9 & 1.6 & 2.5 & 3.8 & 2.9 & 3.6 & 4.4 & 3.8 & 4.6 & 5.2 & 2.0 & 2.8 & 3.3 & 1.7 & 2.9 & 3.5 \\
DA & 5.7 & 0.9 & 1.8 & 2.6 & 1.1 & 2.2 & 3.2 & 2.4 & 3.1 & 3.7 & 3.2 & 4.2 & 4.8 & 1.7 & 2.5 & 3.1 & 1.2 & 2.4 & 3.3 \\
GT & 5.4 & 1.0 & 1.9 & 2.7 & 1.3 & 2.1 & 3.5 & 2.6 & 3.5 & 4.2 & 4.0 & 4.8 & 5.3 & 1.9 & 2.7 & 3.4 & 1.5 & 2.6 & 3.6 \\
\midrule
\multicolumn{20}{c}{\it GRE} \\
\midrule
TC & 4.7 & 0.4 & 0.8 & 1.2 & 0.5 & 0.9 & 1.3 & 0.8 & 1.2 & 2.0 & 3.1 & 3.7 & 4.9 & 1.1 & 1.6 & 2.1 & 0.7 & 1.5 & 2.2 \\
SE & 5.0 & 0.3 & 0.7 & 1.0 & 0.4 & 0.8 & 1.1 & 1.2 & 1.9 & 2.8 & 3.5 & 4.2 & 4.6 & 0.9 & 1.3 & 1.8 & 0.8 & 1.4 & 2.1 \\
RC & 5.6 & 0.8 & 1.3 & 1.9 & 1.0 & 1.5 & 2.2 & 2.5 & 3.4 & 4.0 & 4.1 & 4.7 & 5.2 & 1.4 & 2.0 & 2.6 & 1.1 & 1.9 & 2.5 \\
QC & 6.0 & 1.1 & 1.7 & 2.4 & 1.4 & 2.0 & 2.7 & 2.7 & 3.6 & 4.3 & 3.9 & 4.7 & 5.1 & 1.7 & 2.3 & 3.0 & 1.4 & 2.1 & 2.9 \\
NE & 8.2 & 1.0 & 1.6 & 2.2 & 1.2 & 1.9 & 2.5 & 2.8 & 3.5 & 4.2 & 4.0 & 4.6 & 5.0 & 1.6 & 2.2 & 2.8 & 1.3 & 2.0 & 2.7 \\
DI & 6.2 & 1.3 & 1.9 & 2.8 & 1.5 & 2.2 & 3.1 & 3.1 & 4.0 & 4.6 & 4.2 & 4.9 & 5.4 & 1.9 & 2.6 & 3.4 & 1.6 & 2.4 & 3.2 \\
\midrule
\multicolumn{20}{c}{\it GMAT} \\
\midrule
CR & 4.9 & 0.5 & 0.9 & 1.3 & 0.6 & 1.0 & 1.4 & 1.4 & 2.0 & 2.7 & 3.2 & 3.8 & 4.5 & 1.0 & 1.6 & 2.2 & 0.8 & 1.5 & 2.0 \\
RC & 5.2 & 0.7 & 1.2 & 1.6 & 0.9 & 1.3 & 1.9 & 2.0 & 2.8 & 3.5 & 3.6 & 4.3 & 5.0 & 1.3 & 1.8 & 2.4 & 1.0 & 1.7 & 2.3 \\
PS & 6.3 & 1.4 & 2.1 & 2.7 & 1.8 & 2.5 & 3.3 & 3.0 & 3.9 & 4.6 & 4.1 & 4.9 & 5.3 & 2.0 & 2.7 & 3.5 & 1.6 & 2.3 & 3.1 \\
DS & 6.1 & 1.3 & 2.0 & 2.6 & 1.7 & 2.4 & 3.0 & 2.9 & 3.8 & 4.4 & 4.0 & 4.7 & 5.2 & 1.8 & 2.5 & 3.2 & 1.5 & 2.2 & 2.9 \\
IR & 6.5 & 1.5 & 2.2 & 3.0 & 1.9 & 2.6 & 3.5 & 3.2 & 4.1 & 4.8 & 4.3 & 5.0 & 5.4 & 2.1 & 2.9 & 3.7 & 1.7 & 2.5 & 3.4 \\
\midrule
\multicolumn{20}{c}{\it TOEFL} \\
\midrule
FI & 5.5 & 0.4 & 0.7 & 1.0 & 0.5 & 0.9 & 1.2 & 0.9 & 1.3 & 1.9 & 1.7 & 2.3 & 2.9 & 0.8 & 1.1 & 1.5 & 0.6 & 1.0 & 1.6 \\
IR & 5.8 & 0.8 & 1.1 & 1.5 & 1.0 & 1.4 & 1.9 & 1.6 & 2.1 & 2.7 & 2.4 & 3.1 & 3.8 & 1.2 & 1.6 & 2.0 & 1.1 & 1.5 & 2.2 \\
TS & 5.1 & 0.5 & 0.8 & 1.2 & 0.6 & 1.0 & 1.4 & 1.2 & 1.7 & 2.3 & 1.9 & 2.5 & 3.4 & 0.9 & 1.3 & 1.8 & 0.7 & 1.2 & 1.6 \\
FI & 0.6 & 0.6 & 0.9 & 1.3 & 0.7 & 1.1 & 1.6 & 1.5 & 2.0 & 2.6 & 2.1 & 2.8 & 3.9 & 1.1 & 1.5 & 2.0 & 0.9 & 1.4 & 1.9 \\
IF & 2.7 & 0.9 & 1.3 & 1.8 & 1.2 & 1.7 & 2.3 & 1.9 & 2.6 & 3.2 & 2.6 & 3.4 & 4.7 & 1.4 & 1.9 & 2.6 & 1.2 & 1.8 & 2.5 \\
\midrule
\multicolumn{20}{c}{\it IELTS} \\
\midrule
II & 5.4 & 0.5 & 0.7 & 1.0 & 0.6 & 1.0 & 1.4 & 1.3 & 1.8 & 2.5 & 2.0 & 2.7 & 3.6 & 0.9 & 1.3 & 1.7 & 0.7 & 1.1 & 1.5 \\
MS & 5.7 & 0.7 & 1.0 & 1.3 & 0.9 & 1.3 & 1.8 & 1.7 & 2.2 & 2.9 & 2.5 & 3.2 & 4.1 & 1.1 & 1.5 & 2.0 & 0.8 & 1.2 & 1.7 \\
CP & 5.3 & 0.6 & 0.9 & 1.2 & 0.7 & 1.2 & 1.7 & 1.6 & 2.1 & 2.8 & 2.3 & 3.0 & 4.0 & 1.0 & 1.4 & 1.9 & 0.9 & 1.3 & 1.8 \\
IM & 5.8 & 0.8 & 1.1 & 1.5 & 1.0 & 1.5 & 2.1 & 2.0 & 2.7 & 3.4 & 2.7 & 3.5 & 4.6 & 1.2 & 1.7 & 2.2 & 1.1 & 1.6 & 2.1 \\
CL & 6.2 & 1.0 & 1.4 & 2.0 & 1.2 & 1.8 & 2.5 & 2.3 & 3.0 & 3.8 & 3.0 & 3.9 & 5.0 & 1.4 & 2.0 & 2.7 & 1.2 & 1.7 & 2.3 \\
SA & 5.5 & 0.7 & 1.0 & 1.4 & 0.8 & 1.3 & 1.9 & 1.5 & 2.0 & 2.6 & 2.2 & 2.9 & 4.2 & 1.1 & 1.5 & 2.1 & 0.9 & 1.4 & 1.8 \\
\bottomrule
\end{tabular}}
\label{tab:main-expt-std}
\end{table*}

Table \ref{tab:main-expt-std} complements with Table \ref{tab:main-expt} with standard deviations.

\subsection{Additional Analyses of Modality-Induced Failures}
\label{app:rq1-modality}

A closer examination of multimodal EST questions reveals several recurring failure modes that cut across models and prompting strategies:

First, we find that {\bf misalignment errors} dominate in tasks requiring table–text or text–figure integration. Models frequently retrieve the correct local evidence (e.g., a row or column from a table) but then conflate it with irrelevant contextual information, producing internally coherent but incorrect rationales. Unlike humans, who naturally ground their reasoning in visual scanning and cross-referencing, LLMs rely on implicit token co-occurrence patterns, which are brittle under distribution shifts in layout or labeling.

Second, {\bf arithmetic and normalization mistakes} emerge when quantitative reasoning spans modalities. In GRE Data Interpretation, for instance, models can identify the relevant chart element but fail to convert absolute differences into relative percentages, leading to incorrect comparative judgments. These failures suggest weaknesses in bridging symbolic numeric operations with natural language descriptions, particularly when multiple units, scales, or denominators must be tracked simultaneously.

Third, {\bf over-trust in salient cues} is a pervasive issue. When figures or diagrams contain visually prominent but logically irrelevant elements (e.g., a bolded number or a large bar in a chart), models often anchor on these features even when the question explicitly requires a subtler comparison. Humans, by contrast, employ metacognitive checks such as rereading the question stem to confirm task requirements.

Finally, we observe {\bf compounding variance across modalities}. Errors often cascade: a misread in the textual description can propagate into the tabular lookup, which then interacts with an arithmetic miscalculation, producing errors that appear systematic but in fact result from small deviations at multiple stages. This multi-stage fragility highlights the gap between current LLMs’ sequential token prediction and the hierarchical integration that multimodal reasoning demands.

\textbf{Insights.} These analyses underscore that modality complexity introduces qualitatively new challenges beyond scaling model size or training data. Future work on EST-style problem solving must therefore move beyond token-level modeling to incorporate explicit alignment, symbolic grounding, and verification mechanisms that can emulate the multi-channel reasoning strategies of human test-takers.

\subsection{ LLMs vs. Human Testers}
\label{app:expt-human}

\begin{table}[h]
\centering
\caption{Statistical significance of performance differences between human testers and best-performing LLM (McNemar's test, $p$-values). Asterisks indicate significance levels: *** $p < 0.001$, ** $p < 0.01$, * $p < 0.05$, ns: not significant.}
\label{tab:significance}
\resizebox{\columnwidth}{!}{
\begin{tabular}{llcc}
\toprule
\textbf{Exam} &  \textbf{Task} &  \textbf{Best LLM} &  \textbf{$p$-value} \\
\midrule
\multirow{7}{*}{SAT} 
& II &  Gemini-2.5 (ToT) &  0.001*** \\
& CS &  Gemini-2.5 (CoT) &  0.089ns \\
& EI &  GPT-5 (CoT) &  0.023* \\
& EC &  GPT-5 (CoT) &  0.156ns \\
& AG &  Gemini-2.5 (ToT) &  <0.001*** \\
& DA &  GPT-5 (ToT) &  <0.001*** \\
& GT &  GPT-5 (CoT) &  0.008** \\
\midrule
\multirow{6}{*}{GRE}
& TC &  GPT-5 (ToT) &  0.012* \\
& SE &  GPT-5 (CoT) &  0.034* \\
& RC &  GPT-5 (CoT) &  0.003** \\
& QC &  GPT-5 (ToT) &  <0.001*** \\
& NE &  Human superior &  <0.001*** \\
& DI &  Human superior &  <0.001*** \\
\midrule
\multirow{5}{*}{GMAT}
& CR &  Claude-S4 (ToT) &  0.019* \\
& RC &  Claude-S4 (ToT) &  0.005** \\
& PS &  Human superior &  <0.001*** \\
& DS &  Human superior &  <0.001*** \\
& IR &  Human superior &  <0.001*** \\
\midrule
\multirow{5}{*}{TOEFL}
& FI &  GPT-5 (ToT) &  0.002** \\
& IR &  GPT-5 (CoT) &  0.004** \\
& TS &  Human superior &  0.178ns \\
& FI (Listen) &  GPT-5 (ToT) &  0.028* \\
& IF &  GPT-5 (ToT) &  <0.001*** \\
\midrule
\multirow{6}{*}{IELTS}
& II &  GPT-5 (ToT) &  0.041* \\
& MS &  Human superior &  0.015* \\
& CP &  Claude-S4 (ToT) &  <0.001*** \\
& IM &  GPT-5 (ToT) &  0.092ns \\
& CL &  GPT-5 (ToT) &  0.006** \\
& SA &  Human superior &  0.134ns \\
\bottomrule
\end{tabular}}
\end{table}

Across tasks and modalities, we observe qualitatively different patterns of variability between human testers and LLMs, with significance measures shown in Table \ref{tab:significance}. Note that human variability is driven primarily by background knowledge, test-taking habits, fatigue, and individual strategy preferences; mistakes tend to be idiosyncratic and cluster by prior exposure (e.g., comfort with specific grammar rules or math subskills). 

In contrast, LLM variability is shaped by decoding stochasticity, prompt sensitivity, and fragile intermediate reasoning: the same model can oscillate between correct and incorrect answers when minor surface features change (instruction phrasing, option order, or distractor salience). 

Humans often adapt strategy mid-session and exhibit metacognitive checks (skimming, re-reading, sanity checks on units or logic), whereas LLMs more frequently display ``local optimum'' traps (e.g., latching onto a salient but irrelevant cue) or instruction-following drift without self-correction. Variability is also modality-dependent: humans degrade with cognitive load and time pressure, while LLMs degrade more when cross-representation alignment is required (text–table, text–image, text–audio), reflecting weaknesses in binding and content grounding rather than domain knowledge alone.

\subsection{Impact by Prompting Complexity} 
\label{app:expt-prompt-complexity}

We also find that more sophisticated prompting strategies (e.g., ToT) do not consistently lead to better performance, although more enriched reasoning is provided:

\begin{theobox}
{\bf Case Study} (Text Completion):
{\it Although it is easy to imagine that the \rule{0.8cm}{0.15mm} of technological innovation has accelerated ... ... innovation has proceeded at a fairly \rule{0.8cm}{0.15mm} pace since the Industrial Revolution. Options: 1. (i) tempo, (ii) constant. Options: 2. (i) novelty, (ii) sporadic. Options: 3. (i) velocity, (ii) erratic.}

\vspace{3pt}
{\bf CoT focuses on overall sentence coherence:} The sentence suggests a contrast between the perception that innovation has {\it accelerated} and the ... ... Thus, the correct answer is {\bf Option 1.}

\vspace{3pt}
{\bf ToT forces blank-by-blank exploration:} Branch 1: For first blank. (1.a) Option "tempo" → meaning = speed. (1.b) Option "novelty" → meaning = newness ... ... Branch 2: For the second blank (2.a) Option "constant" → meaning = unchanging ... ... {\bf Final Answer: Option 3 (i) velocity, (ii) erratic.} (LLM gets confused due to multiple branches and partial fits.)
\end{theobox}

This suggests that complex reasoning frameworks may sometimes introduce additional cognitive overhead without corresponding gains in accuracy, particularly for models not explicitly optimized for such structured inference. 

We further find that LLM performance may also be influenced by prompting and decoding choices, with different trends:

(1) {\bf Chain-of-thought (CoT)} generally regularizes reasoning on verbal items by externalizing intermediate structure, but it can also amplify spurious rationales when the initial trajectory is off-distribution. This is particularly visible in GRE Sentence Equivalence, where once the model locks onto a semantically plausible but incorrect synonym, subsequent steps reinforce the error rather than revising it. In math-heavy tasks like SAT Algebra, CoT sometimes leads to over-elaboration, generating unnecessary symbolic steps that increase the chance of arithmetic drift.

(2) {\bf Tree-of-thought (ToT)} tends to help on search-like or data-integration tasks, yet it introduces longer reasoning paths that sometimes accumulate small local errors or trigger premature branch pruning. For instance, in GMAT Data Sufficiency, ToT can improve systematic exploration of conditions but is also prone to “path explosion,” where irrelevant branches dominate and obscure the correct constraint check. Similarly, in GRE Data Interpretation, ToT may spread reasoning across multiple chart elements without recombining them, leading to fragmented conclusions.

(3) {\bf In-context learning (ICL)} works best when exemplars match the target item’s latent schema (discourse function, syntactic frame, or quantitative template); schema-mismatched exemplars can anchor the model to the wrong solution space. In IELTS Matching Sentence tasks, schema-aligned exemplars guide the model toward identifying discourse relations effectively, whereas mismatched exemplars bias the model toward surface string overlaps. In TOEFL Inference questions, exemplar mismatch often causes the model to ignore pragmatic cues like tone or implied stance, overfitting instead to lexical similarity.

\textbf{Insights.} These observations suggest that elicitation strategies interact strongly with task type and associated failure modes. CoT excels in tasks requiring layered linguistic reasoning but exacerbates semantic anchoring errors when the first step is flawed. ToT adds value where systematic exploration is necessary (tables, condition checks, multi-source reasoning), but it magnifies variance when intermediate steps are noisy or poorly pruned. ICL is powerful when schema alignment is possible, but fragile when exposed to distributional mismatch between exemplars and target questions. Together, these findings underscore that reliable EST problem solving requires not only robust prompt design but also adaptive elicitation strategies that are sensitive to the structural demands and common pitfalls of each task family.

\subsection{Influence of Question Difficulty}
\label{app:expt-difficulty}

\begin{figure*}[t]
    \centering
    \includegraphics[width=\textwidth]{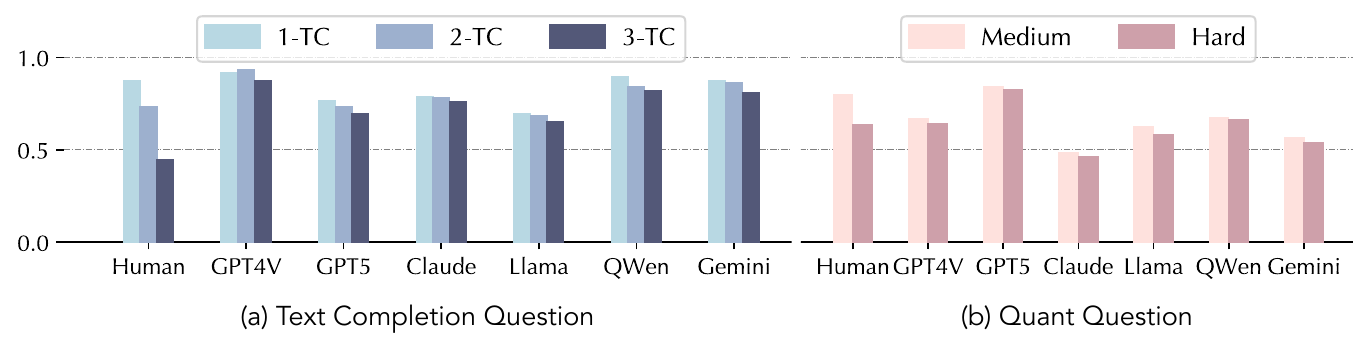}
    \caption{LLM performance across varying levels of question difficulty, using CoT due to its representativeness. We focus on GRE text completion tasks with 1-, 2-, and 3-blanks, as well as available medium- and hard-level quantitative problems.}
    \label{fig:expt-difficulty}
\end{figure*}

The question types in Table~\ref{tab:main-expt} already reflect varying levels of difficulty—for example, inference questions in TOEFL reading are generally more challenging than factual information questions. 

We further investigate how question difficulty influences LLM performance. We evaluate on GRE, which allow for clearer categorization, wherein text completion (TC) questions are divided into one-, two-, and three-blank formats with a greater number of blanks corresponds to higher difficulty. Similarly, quantitative (Quant) questions are pre-labeled as either medium or hard. Figure~\ref{fig:expt-difficulty} presents LLM performance across these difficulty levels. Interestingly, we observe no clear performance degradation as difficulty increases, where human testers show a significant decline in answer correctness as the difficulty increases. These results suggest that LLMs may not be sensitive to human-defined difficulty levels and instead exhibit an equilibrium across structurally similar problems, regardless of their intended complexity in English or mathematical settings.

\textbf{Note on Difficulty Categorization:} Difficulty levels in Figure \ref{fig:expt-difficulty} use official ETS metadata. GRE Quantitative questions are labeled as ``medium'' or ``hard'' in source materials; Text Completion difficulty corresponds to the number of blanks (1/2/3), a structural feature standardized in test design that correlates with cognitive complexity. We do not impose difficulty labels on exams (SAT, GMAT, TOEFL, IELTS) lacking official metadata.

\subsection{Inference Efficiency}
\label{app:expt-time}

\begin{figure*}[t]
    \centering
    \includegraphics[width=\textwidth]{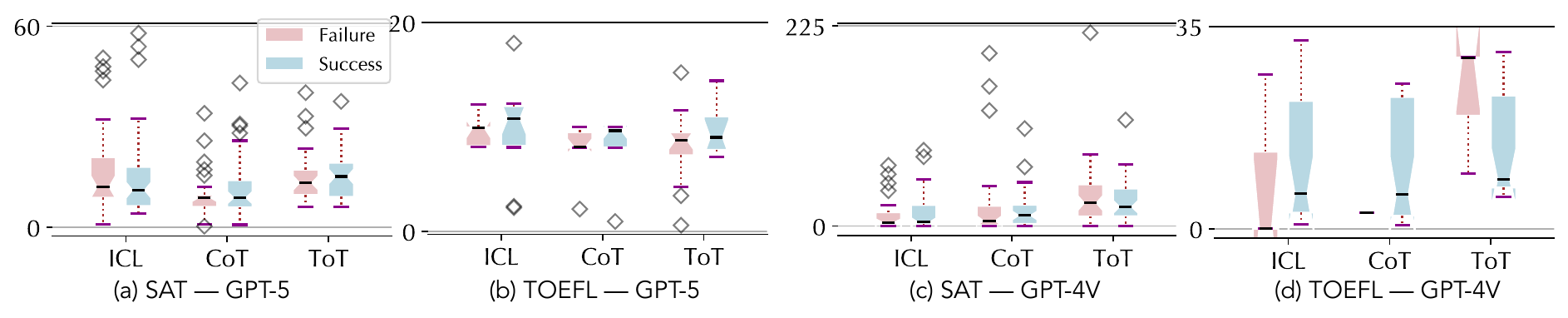}
    \caption{Inference time (in seconds) for failed and successful cases. More results are in Figure \ref{fig:expt-time-2}.}
    \label{fig:expt-time}
\end{figure*}

Another important consideration is the inference time of LLMs. To analyze the relationship between inference time and answer correctness, we record the generation time (in seconds) for each response and categorize the results into two groups: correctly answered and incorrectly answered questions. We then plot the distribution of inference times for both groups, as shown in Figure~\ref{fig:expt-time}.

From the box plots, we observe that the inference times for correct and incorrect predictions are similar, without a notable separation between the two groups. To statistically validate this observation, we perform a two-tailed Mann–Whitney U test \citep{mcknight2010mann}. The Mann–Whitney U 
\begin{table}[t]
\centering
\small
\caption{Mann--Whitney U test of the inference time between failed and successful cases. We report p-values to assess the statistical significance of differences between the success and failure groups. Additional results are presented in Table~\ref{tab:p-value-2}.}
\resizebox{\linewidth}{!}{
\begin{tabular}{ccccccc}
\toprule
\multicolumn{1}{c}{\multirow{2}{*}{Exam}} &
\multicolumn{3}{c}{GPT-5} &
\multicolumn{3}{c}{GPT-4V} \\
\cmidrule(lr){2-4}\cmidrule(lr){5-7}
& ICL & CoT & ToT & ICL & CoT & ToT \\
\midrule
SAT   & 0.278 & 0.814 & 0.443 & 0.197 & 0.117 & 0.512 \\
TOEFL & 0.610 & 0.389 & 0.515 & 0.295 & 0.640 & 0.115 \\
\bottomrule
\end{tabular}}
\label{tab:p-value}
\end{table}
test is a non-parametric hypothesis test that assesses whether two independent samples come from the same distribution. It evaluates whether the distributions differ in location (median) or overall shape. As listed in Table \ref{tab:p-value}, across all evaluated models, the Mann–Whitney U tests yield  p-values higher than 0.05 (a commonly used significance level), indicating no statistically significant difference between the inference time distributions of correct and incorrect predictions. This suggests that the time an LLM spends on answering a question does not correlate with answer correctness. Inference time appears largely independent of answer quality on the EST benchmark.

\begin{figure*}[t]
    \centering
    \includegraphics[width=\textwidth]{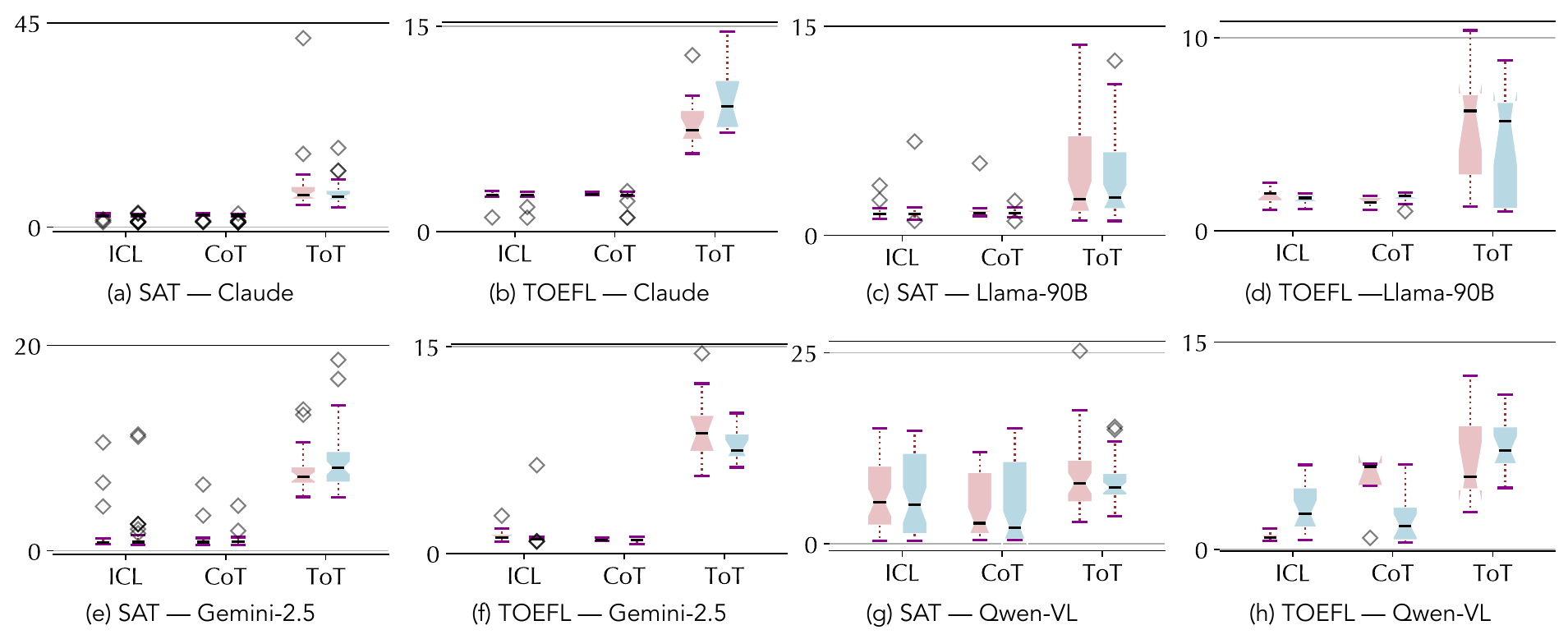}
    \caption{Inference time (in seconds) for failed and successful cases. Complement to Figure \ref{fig:expt-time}.}
    \label{fig:expt-time-2}
\end{figure*}

Figure \ref{fig:expt-time-2} and Table \ref{tab:p-value-2} provides more experimental results for inference time across success and failure cases. 

\begin{table*}[h]
\centering
\small
\caption{Mann–Whitney U test of the inference time between failed and successful cases. Complement to Table \ref{tab:p-value}.}
\scalebox{0.9}{
\begin{tabular}{ccccccccccccc}
\toprule
\multicolumn{1}{c}{\multirow{2}{*}{Exam}} &
\multicolumn{3}{c}{Claude-Sonnet-4} &
\multicolumn{3}{c}{Llama-4-Scout-17B} & \multicolumn{3}{c}{Gemini-2.5} &
\multicolumn{3}{c}{Qwen-VL-Max} \\
\cmidrule(lr){2-4}\cmidrule(lr){5-7}\cmidrule(lr){8-10}\cmidrule(lr){11-13}\
& ICL & CoT & ToT & ICL & CoT & ToT & ICL & CoT & ToT & ICL & CoT & ToT \\
\midrule
SAT  & 0.572 & 0.359 & 0.336 & 0.786 & 0.817 & 0.619 & 0.994 & 0.449 & 0.105 & 0.734 & 0.377 & 0.884 \\
TOEFL & 0.903 & 0.137 & 0.084 &  0.449 & 0.084 & 0.360 & 0.267 & 0.414 & 0.231 & 0.159 & 0.088 & 0.374 \\
\bottomrule
\end{tabular}}
\label{tab:p-value-2}
\end{table*}

\section{Additional Experimental Results and Discussions for RQ$_2$}
\label{app:expt-more-rq2}

\subsection{Evaluation Metrics for Cognitive Trajectories}
\label{app:breakdown_metrics}

\begin{table*}[!t]
\centering
\caption{Evaluation metrics applied to individual Cognitive DAG nodes across the three pedagogical categories in \bench. The heterogeneous nature of the nodes (ranging from text extraction to mathematical formulation) necessitates a diverse suite of deterministic and semantic metrics.}
\label{tab:breakdown-metrics}
\resizebox{\textwidth}{!}{
\begin{tabular}{lllp{6.5cm}}
\toprule
\textbf{Pedagogical Category} & \textbf{Cognitive DAG Node} & \textbf{Evaluation Metric} & \textbf{Diagnostic Justification} \\
\midrule
\multirow{6}{*}{\shortstack[l]{\textbf{Category I:}\\\textbf{Knowledge-Intensive}\\\textbf{Retrieval}}} 
& $c_{\text{syntax\_parse}}$ / $c_{\text{semantic\_parse}}$ & Exact Match (Accuracy) & Verifies structural or semantic classification (e.g., identifying sentence tense or missing part of speech). \\
& $c_{\text{rule\_match}}$ / $c_{\text{logical\_scope}}$ & Token-level IoU / F1 & Measures the overlap between the LLM's identified logical anchor (e.g., contrast words) and the ground truth. \\
& $c_{\text{predict}}$ / $c_{\text{resolve}}$ & BERTScore & Evaluates the semantic plausibility of the predicted word or phrase, accepting valid paraphrases. \\
\midrule
\multirow{6}{*}{\shortstack[l]{\textbf{Category II:}\\\textbf{Reasoning-Intensive}\\\textbf{Execution}}} 
& $c_{\text{analytical\_goal}}$ / $c_{\text{model}}$ & Exact Match (Accuracy) & Classifies if the model correctly identified the math/logic domain (e.g., identifying a problem as ''probability''). \\
& $c_{\text{visual\_parse}}$ / $c_{\text{formulate\_eq}}$ & Symbolic Equivalency & Uses a CAS (e.g., SymPy) or structural matching to verify formulas, ignoring surface-level formatting ($x+y$ vs. $y+x$). \\
& $c_{\text{analyze}}$ / $c_{\text{compute}}$ & $1 - \text{Normalized RMSE}$ & Evaluates numeric execution, providing partial credit for minor rounding differences while penalizing gross arithmetic failures. \\
\midrule
\multirow{6}{*}{\shortstack[l]{\textbf{Category III:}\\\textbf{Hybrid}\\\textbf{Integration}}} 
& $c_{\text{intent}}$ / $c_{\text{entities}}$ & Exact Match (Accuracy) & Strictly evaluates whether the correct target subject or comparative entities were isolated from the prompt. \\
& $c_{\text{perceive\_text}}$ / $c_{\text{constraints}}$ & Sentence-level IoU & Assesses the accuracy of localizing the specific evidence spans or logical constraints within a dense passage/audio. \\
& $c_{\text{extract}}$ / $c_{\text{evaluate}}$ & BERTScore & Measures the fidelity of the final deductive reasoning or summarized evidence against the ground-truth rationale. \\
\bottomrule
\end{tabular}
}
\end{table*}

Because the cognitive trajectory models a heterogeneous problem-solving process—ranging from finding a specific sentence in a reading passage to computing a complex polynomial—no single metric can adequately evaluate all intermediate reasoning steps. Therefore, as detailed in Table \ref{tab:breakdown-metrics}, we deploy a suite of tailored metrics depending on the cognitive function of the cognitive step.

\textbf{Extractive and Localization Nodes (IoU \& F1):} For nodes that require the LLM to scan and isolate information from a larger context (e.g., $c_{\text{logical\_scope}}$ in Category I, or $c_{\text{perceive\_text}}$ in Category II), we utilize Intersection over Union (IoU) and Token-level F1 scores. These metrics are pedagogically appropriate because they measure \textit{attention precision}. If a model highlights an entire paragraph when the evidence is contained in a single sentence, the IoU naturally penalizes the model for a lack of focus, reflecting a common human test-taking inefficiency.

\textbf{Mathematical and Execution Nodes (Symbolic Equivalency \& RMSE):} Category II nodes demand rigorous computational precision. Evaluating intermediate mathematical formulation ($c_{\text{formulate\_eq}}$) via standard string matching frequently yields false negatives (e.g., penalizing $2x + y = 10$ against $y = 10 - 2x$). To resolve this, we utilize a Computer Algebra System (CAS), such as SymPy, to verify \textit{Symbolic Equivalency}. For the final execution node ($c_{\text{compute}}$), we employ $1 - \text{Normalized RMSE}$ for continuous numeric outputs, which strictly penalizes fundamental arithmetic hallucinations while gracefully handling minor floating-point or rounding deviations.

\textbf{Generative and Deductive Nodes (BERTScore):} For nodes requiring the synthesis of meaning or final logical resolution (e.g., $c_{\text{resolve}}$ or $c_{\text{extract}}$), strict exact-match metrics are overly punitive. An LLM might correctly infer a concept but express it using slightly different vocabulary than the ground truth. To accommodate valid linguistic variations while ensuring the core pedagogical reasoning remains intact, we employ BERTScore to evaluate the semantic fidelity of the generated text against the official standardized test explanations.

\subsection{Complementing experimental results}
\label{app:add-expt-rq2}

Figure \ref{fig:expt-breakdown-gpt4o} and \ref{fig:expt-breakdown-gemini} provides additional breakdown analysis on other LLMs, wherein the observation aligns with Section \ref{ssec:expt-breakdown}.

\begin{figure*}[h]
    \centering
    \includegraphics[width=\textwidth]{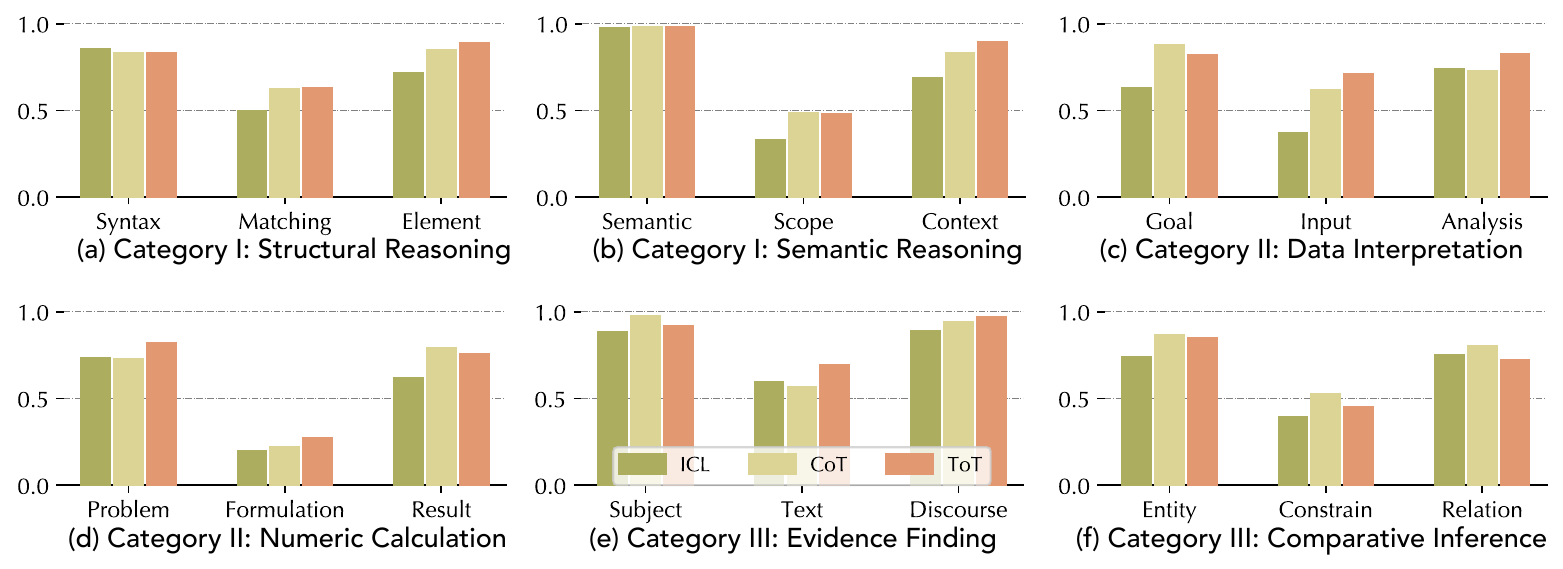}
    \caption{Breakdown analysis on Claude.}
    \label{fig:expt-breakdown}
\end{figure*}

\begin{figure*}[h]
    \centering
    \includegraphics[width=\textwidth]{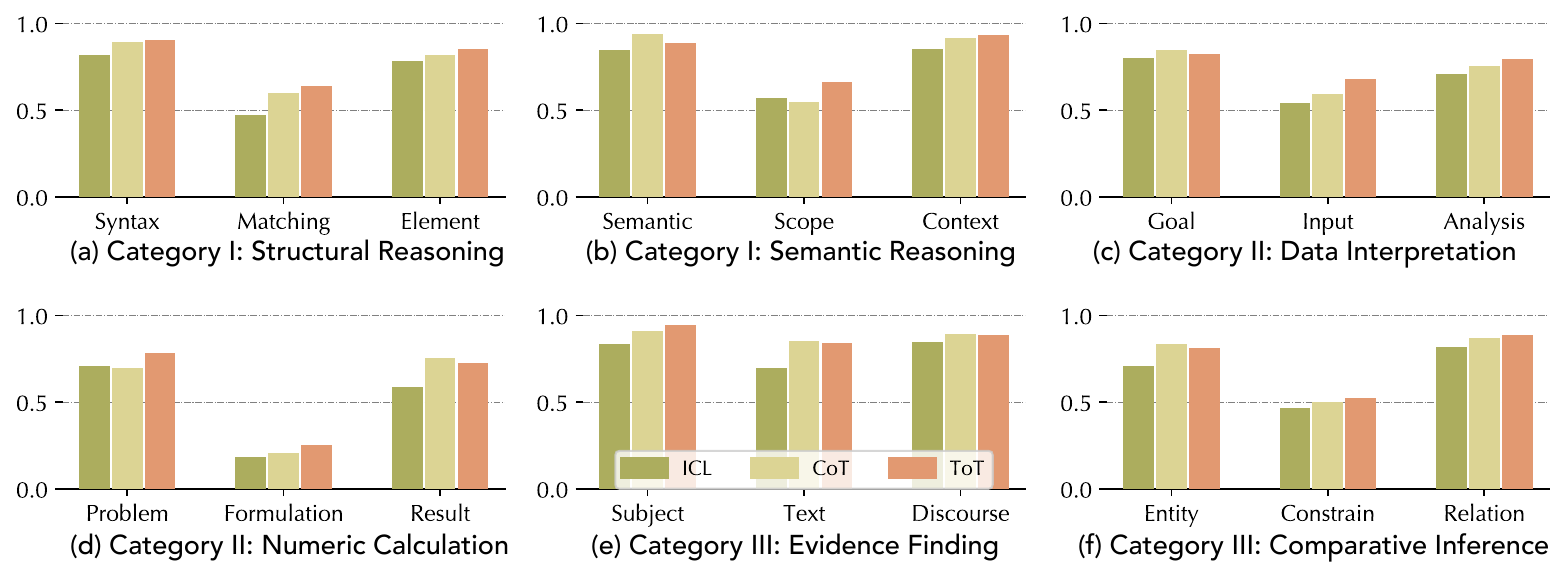}
    \caption{Breakdown analysis on Gemini.}
    \label{fig:expt-breakdown-gemini}
\end{figure*}

\subsection{Additional Case Studies}
\label{app:more-case-study}

This section provides additional case studies complement to observations and conclusions in RQ$_3$.

\begin{theobox}
{\bf Case Study (GRE Quant – Numeric Entry):}
{\it "If the sum of three consecutive odd integers is 111, what is the smallest of the three?"}

\vspace{3pt}
{\bf LLM Reasoning:} Claude generates an incorrect expression: $x + x+1 + x+2 = 111$ as it treats the numbers as consecutive integers rather than odd integers (which should be $x + x+2 + x+4 = 111$). 
\end{theobox}

{\bf Interpretation: Numeric entry and multi-modality significantly impede LLM reasoning.} Tasks involving numeric input and multimodal understanding (e.g., math from SAT) remain particularly challenging for LLMs. Unlike classification-style questions with fixed answer choices, numeric-entry tasks require precise mathematical formulation, symbolic manipulation, and error-free calculation—all of which are error-prone in current models. 

The challenge is amplified in multimodal settings, where the model must align visual, tabular, or symbolic inputs with textual queries before reasoning can even begin.

\begin{theobox}
{\bf Case Study (GMAT Integrated Reasoning – Table + Math Computation):}
A question requires {\it "selecting a product with the highest profit margin based on a table of costs and revenues."} GPT-4V incorrectly reads the table and subtracts the cost from total units sold rather than revenue, leading to an invalid numeric result. 
\end{theobox}

\begin{theobox}
{\bf Case Study (GRE – Semantic Reasoning):}
{\it "Select the pair of words that best completes the sentence: 'While the professor’s tone was ostensibly \_\_\_\_, her critique was undeniably severe and cutting.'"}  
Options:  
(A) respectful – insulting  
(B) conciliatory – harsh  
(C) disinterested – involved  
...

\vspace{3pt}
{\bf LLM Reasoning:}  
GPT-5 selects (A) due to surface-level antonymy (“respectful” vs. “insulting”), but fails to resolve the nuanced implication of “ostensibly” versus “undeniably,” which is essential for semantic disambiguation. Claude performs similarly, missing the contrastive logic implied by adverbs. Only Gemini-Pro correctly identifies (B), recognizing the indirect semantic contrast.
\end{theobox}

{\bf Interpretation:} This illustrates how LLMs, despite strong lexical capabilities, still struggle with subtle discourse-level signals that guide meaning, such as modal adverbs or pragmatic contrast. It reinforces your claim that deeper reasoning (not surface matching) is the primary challenge.

\begin{theobox}
{\bf Case Study (IELTS Listening – Evidence Localization):}  
{\it "What is the speaker's main reason for supporting the expansion of the city park?"}  
Audio clip: The speaker describes multiple benefits of expanding a city park, including noise reduction, community wellness, and increased biodiversity.

\vspace{3pt}
{\bf LLM Reasoning:}  
Using Whisper-transcribed audio, Qwen and GPT-4V highlight “noise reduction” as the answer because it is mentioned first and most clearly. However, the correct answer is “community wellness,” which is emphasized later in the speech with supporting elaboration. Only Gemini-Pro correctly weighs the relative emphasis across the transcript.
\end{theobox}

{\bf Interpretation:} This example shows that current models tend to over-prioritize the first mentioned or most literal content in a multimodal context, and fail to simulate human-like discourse prioritization. It also suggests weaknesses in aligning Whisper transcripts with reasoning modules.

\begin{theobox}
{\bf Case Study (SAT Reading – Evidence Pairing):}  
{\it "Which of the following best supports the answer to the previous question?"}  
Passage: A student challenges the conclusions of a scientific article.  
Main question: “Which claim does the student most strongly refute?”  
Evidence question: “Which line best supports that refutation?”

\vspace{3pt}
{\bf LLM Reasoning:}  
Claude selects a sentence that contains a general critique but does not directly support the earlier answer. GPT-5 does better at matching tone but fails to anchor the evidence to the refuted claim. Only Gemini-Pro correctly links the reasoning across both questions.  
\end{theobox}
{\bf Breakdown Challenge:  Evidence Localization.} Highlights LLMs’ difficulty in chaining answers across linked questions, especially when reasoning must remain consistent.

\begin{theobox}
{\bf Case Study (GRE Verbal – Logical Structure):}  
{\it "Which of the following best describes the structure of the passage?"}  
Passage: An author introduces a phenomenon, critiques one explanation, and then proposes an alternative.

\vspace{3pt}
{\bf LLM Reasoning:}  
LLaMA-3 and Qwen select options that only capture the first half of the structure (e.g., critique). GPT-4V overgeneralizes to a "compare-and-contrast" structure. Only Claude correctly recognizes the structure as “Introduction → Criticism → Alternative Explanation.”  

\end{theobox}
{\bf Breakdown Challenge: Structural Reasoning.} Illustrates that models struggle to track abstract rhetorical moves across a passage, even when comprehension is accurate.

\begin{theobox}
{\bf Case Study (GMAT Integrated Reasoning – Two-Part Analysis):}  
{\it "Select one answer for each of the following two conditions: (1) Which project has the highest ROI? (2) Which project has the lowest risk?"}  
Tabular data includes five projects with ROI and risk indicators.

\vspace{3pt}
{\bf LLM Reasoning:}  
GPT-4V selects Project C for both ROI and risk, confusing “least cost” with “least risk.” Claude selects correctly for ROI but fails to interpret qualitative risk descriptors. Gemini-Pro and GPT-5 complete both selections correctly.  

\end{theobox}
{\bf Breakdown Challenge: Data Interpretation. } Shows difficulty in multi-constraint reasoning and mapping discrete table fields to textual decision logic.

\begin{theobox}
{\bf Case Study (IELTS Writing – Grammatical Error Correction):}  
{\it "Identify and correct the grammatical error in the following sentence:  
'If she would have gone to the meeting, she could had contributed valuable insight.'"}

\vspace{3pt}
{\bf LLM Reasoning:}  
Qwen changes "she could had" to "she could has," worsening the error. Claude corrects “would have gone” to “had gone” but leaves the second clause unaltered. Only GPT-5 performs both corrections, yielding:  
“If she had gone to the meeting, she could have contributed valuable insight.”  

\end{theobox}

{\bf Breakdown Challenge: Structural Reasoning.} Highlights syntax correction challenges where multiple clauses require coordinated grammatical edits.

\begin{theobox}
{\bf Case Study (GRE Quant – Comparative Judgment):}  
{\it "Quantity A: The square of the average of 3 and 7;  
Quantity B: The average of the squares of 3 and 7."}

\vspace{3pt}
{\bf LLM Reasoning:}  
GPT-4V computes both but incorrectly concludes that Quantity A is greater, mistaking $((3+7)/2)^2 = 25$ as greater than $(3^2 + 7^2)/2 = 29$. LLaMA-3 gives no answer and repeats the prompt. Claude answers correctly but offers no reasoning trace.  

\end{theobox}
{\bf Breakdown Challenge: Comparative Judgment.} Demonstrates common mistakes in applying formulas and comparing expressions under symbolic transformation.

Moreover, we also observe that prompting strategies (ICL, CoT, ToT) do not significantly affect performance in certain stages of breakdown analysis, especially where task complexity is low or answer derivation is mostly local. Below are some case studies to address this:

\begin{theobox}
{\bf Case Study (SAT Reading – Factual Retrieval):}  
{\it "According to the passage, what did the author list as one benefit of urban green space?"}  
The relevant sentence in the passage states: "Green spaces improve air quality and reduce noise levels."

\vspace{3pt}
{\bf LLM Reasoning across Prompts:}  
All three prompting strategies (ICL, CoT, ToT) lead to the same correct output across GPT-5 and Claude. In each case, the models locate the exact supporting sentence and extract "improve air quality" or "reduce noise levels" without variation. CoT and ToT generate unnecessary intermediate steps without improving the final answer.  

\end{theobox}
{\bf Breakdown Relevance:} Evidence Finding (step: locate and extract factual information).
Insight: Prompting complexity doesn't help when the required reasoning is local and unambiguous.

\begin{theobox}
{\bf Case Study (GRE Verbal – 1-Blank Text Completion):}  
{\it "The scientist's explanation was praised for its clarity and \_\_\_\_, making it accessible to a general audience."}  
Options: (A) convolution, (B) transparency, (C) complexity...

\vspace{3pt}
{\bf LLM Reasoning across Prompts:}  
All strategies (ICL, CoT, ToT) result in the selection of (B) "transparency." The reasoning is nearly identical: the model detects positive sentiment from "praised" and "clarity," and eliminates antonymic distractors like "convolution." CoT and ToT elaborate more, but do not change the choice or rationale.  

\end{theobox}
{\bf Breakdown Relevance:} Semantic Reasoning (step: sentiment alignment and elimination).
Insight: For simple semantic alignment, ICL already suffices, and additional reasoning scaffolds don't yield improvement.

\begin{theobox}
{\bf Case Study (GMAT Quant – Basic Arithmetic):}  
{\it "What is the value of $3x + 2$ if $x = 5$?"}

\vspace{3pt}
{\bf LLM Reasoning across Prompts:}  
All prompting strategies produce the correct answer, 17, with or without intermediate steps. CoT redundantly walks through "$3x = 15$, then $15 + 2 = 17$," while ToT splits the steps further into node-like structures. None of the strategies reduce error, latency, or confidence.  

\end{theobox}
{\bf Breakdown Relevance:} Numeric Calculation (step: direct substitution and evaluation).
Insight: When reasoning is shallow and deterministic, prompting scaffolds become unnecessary overhead.

\section{Additional Experimental Results and Discussions for RQ$_3$}
\label{app:expt-more-rq3}

\subsection{Detailed Failure Mode Analysis: The Integration Bottleneck}
\label{app:failure_mode_analysis}

Despite overall accuracy metrics suggesting that modern LLMs are approaching human parity on ESTs, however, our node-level evaluation of distractor susceptibility reveals a consistent, structural vulnerability. Across all tested models, from heavily parameterized frontier models to efficient open-weight architectures, failures are rarely uniformly distributed across the reasoning trajectory. Instead, they exhibit a distinct ``V-shaped'' performance drop, localized almost exclusively at the second node of our three-step cognitive trajectory. We term this the \textbf{Integration Bottleneck}.

As summarized in Table \ref{tab:failure_modes}, models demonstrate high robustness during initial conceptualization (Step 1) and subsequent final-stage execution (Step 3). The catastrophic failures occur when models must bind parsed constraints to specific textual or mathematical representations (Step 2). Generative architectures, optimized for next-token probability, exhibit a strong bias toward plausible rationalization. When presented with a distractor option that contains a \textit{Partial Truth} or a \textit{Faulty Premise}, LLMs tend to hallucinate a valid integration rather than acting as strict logical discriminators.

\begin{table*}[htbp]
\centering
\caption{Qualitative Taxonomy of LLM Failure Modes at the Integration Bottleneck (Step 2). The table maps the specific cognitive vulnerability in each task to the primary distractor trap that successfully derails the model's reasoning trajectory.}
\label{tab:failure_modes}
\resizebox{\textwidth}{!}{
\begin{tabular}{p{3.5cm} p{3.5cm} p{4cm} p{4.5cm}}
\toprule
\textbf{Pedagogical Category} & \textbf{Task \& Bottleneck Node} & \textbf{Primary Distractor Trap} & \textbf{Observed LLM Failure Mechanism} \\
\midrule

\multirow{4}{*}{\shortstack[l]{\textbf{Category I:}\\\textbf{Knowledge-Intensive}\\\textbf{Retrieval}}} 
& \textbf{Structural:} \newline \textit{Match Text} & \textbf{Type III:} Partial Truth \newline (High Lexical Overlap) & Models match the syntactic structure but ignore subtle semantic inversion. They select distractors that ``sound natural'' despite violating local logical scope. \\
\cmidrule{2-4}
& \textbf{Semantic:} \newline \textit{Localize Scope} & \textbf{Type III:} Faulty Premise & Models fail to apply restrictive clauses (e.g., ``except,'' ``unless''). They validate distractors that are true in isolation but false under the specific sentence constraints. \\
\midrule

\multirow{4}{*}{\shortstack[l]{\textbf{Category II:}\\\textbf{Reasoning-Intensive}\\\textbf{Execution}}} 
& \textbf{Data Interpretation:} \newline \textit{Parse Visual Input} & \textbf{Type II / IV:} Out of Scope / Misaligned Variables & Models misalign textual entities with tabular/visual axes. They confidently compute trends using the wrong row/column, failing to reject distractors derived from this exact human error. \\
\cmidrule{2-4}
& \textbf{Numeric Calculation:} \newline \textit{Formulate Math} & \textbf{Type IV:} Execution Error \newline (Flawed Formulation) & The most severe bottleneck. Models correctly model the problem type but fail to translate natural language into a strict algebraic equation, falling for distractors representing incomplete formulations. \\
\midrule

\multirow{4}{*}{\shortstack[l]{\textbf{Category III:}\\\textbf{Hybrid}\\\textbf{Integration}}} 
& \textbf{Evidence Finding:} \newline \textit{Comprehend Text} & \textbf{Type III:} Partial Truth \newline (Paraphrase Hallucination) & Models accept distractors that use exact keywords from the text but combine them to form a conclusion not supported by the actual discourse unit. \\
\cmidrule{2-4}
& \textbf{Comparative Inference:}\newline \textit{Apply Constraints} & \textbf{Type II:} Out of Scope \newline (Unwarranted Assumption) & Models fail to maintain strict logical boundaries. They ``smooth over'' missing information by assuming implicit variables, thereby accepting distractors that require knowledge outside the provided text. \\

\bottomrule
\end{tabular}
}
\end{table*}

\subsubsection*{Category I Vulnerabilities: The Illusion of Fluency}
In Knowledge-Intensive tasks (Syntax and Semantic Reasoning), the primary failure mechanism is an over-reliance on local coherence. Models successfully parse the syntax (Step 1) but fail during \textit{Matching} or \textit{Localizing Scope} (Step 2). Distractor options engineered with high lexical overlap (Type III: Partial Truth) easily deceive the models. Instead of rejecting a syntactically fluent but logically flawed distractor, the LLM generates a post-hoc rationalization to justify why the word ``fits,'' demonstrating an inability to decouple linguistic fluency from rigorous semantic gating.

\subsubsection*{Category II Vulnerabilities: Symbol Grounding Failures}
In Reasoning-Intensive tasks (Data Interpretation and Numeric Calculation), the Integration Bottleneck manifests as a symbol grounding failure. LLMs consistently identify the correct problem type (Step 1), yet experience severe degradation during \textit{Parsing Visual Input} or \textit{Formulating Math} (Step 2). They are highly susceptible to Type IV (Execution) distractors. For instance, if a distractor option represents the answer to an intermediate step of an equation rather than the final solution, the LLM will frequently accept it, losing track of the overarching mathematical constraints established in Step 1.

\subsubsection*{Category III Vulnerabilities: Constraint Relaxation}
For Hybrid Integration tasks (Evidence Finding and Comparative Inference), the bottleneck occurs at \textit{Comprehend Text} and \textit{Apply Constraints}. Here, human-like distractors utilize Type II (Out of Scope) traps by introducing highly plausible but unstated premises. While human test-takers are explicitly trained to identify when information is "Not Given" (e.g., in IELTS Reading or GMAT Data Sufficiency), LLMs struggle with this strict epistemic boundary. The models exhibit ''constraint relaxation,'' implicitly filling in missing data to force a logical connection, thereby erroneously accepting invalid rationales.
\section{Additional Experimental Results and Discussions for RQ$_4$}
\label{app:expt-more-rq4}

In Section \ref{sec:mitigation}, we introduced four bottleneck-specific mitigation strategies designed to repair the reasoning trap (Step 2 of our cognitive trajectory) and reported their impact on a representative subset of \bench. In this appendix, we extend this analysis to a broader range of tasks across all six standardized tests. 

Table \ref{tab:extended_mitigation} presents the comparative performance of baseline Chain-of-Thought (CoT) prompting versus our targeted Mitigated (+Mit.) prompting for GPT-4V, GPT-5, and Claude-Sonnet-4. The baseline scores correspond directly to the CoT columns reported in the main results (Table \ref{tab:main-expt}).

\begin{table*}[htbp]
\caption{Extended evaluation of targeted mitigation strategies across a wider spectrum of \bench tasks. The baseline (Base) scores represent standard Chain-of-Thought (CoT) performance, while Mitigated (+Mit.) scores reflect the application of our bottleneck-specific prompts.}
\label{tab:extended_mitigation}
\centering
\resizebox{0.85\textwidth}{!}{
\begin{tabular}{l cc cc cc}
\toprule
\multirow{2}{*}{\textbf{Task \& Applied Mitigation Strategy}} & \multicolumn{2}{c}{\textbf{GPT-4V}} & \multicolumn{2}{c}{\textbf{GPT-5}} & \multicolumn{2}{c}{\textbf{Claude-S-4}}\\
\cmidrule(lr){2-3} \cmidrule(lr){4-5} \cmidrule(lr){6-7}
& Base & +Mit. & Base & +Mit. & Base & +Mit. \\
\midrule
\rowcolor{gray!10} \multicolumn{7}{l}{\textbf{Strategy 1: Evidence-Anchored CoT (Mitigates Paraphrase Hallucination)}} \\
\textbf{GRE-RC} (Reading Comprehension) & 77.8 & \textbf{93.5} & 87.1 & \textbf{90.4} & 69.3 & \textbf{84.1} \\
\textbf{TOEFL-FI} (Factual Information) & 86.3 & \textbf{91.2} & 93.2 & \textbf{96.5} & 83.9 & \textbf{88.4} \\
\textbf{IELTS-MS} (Matching Sentences) & 85.1 & \textbf{89.4} & 83.0 & \textbf{88.1} & 81.0 & \textbf{85.6} \\

\midrule
\rowcolor{gray!10} \multicolumn{7}{l}{\textbf{Strategy 2: Syntax-First CoT (Mitigates Semantic Drift / Fluency Bias)}} \\
\textbf{GRE-TC} (Text Completion) & 79.5 & \textbf{84.2} & 73.4 & \textbf{78.1} & 61.0 & \textbf{76.7} \\
\textbf{GRE-SE} (Sentence Equivalence) & 81.0 & \textbf{86.3} & 86.5 & \textbf{90.2} & 82.1 & \textbf{87.9} \\
\textbf{SAT-CS} (Cross-Textual Synthesis) & 77.4 & \textbf{83.1} & 82.7 & \textbf{88.5} & 70.2 & \textbf{78.4} \\

\midrule
\rowcolor{gray!10} \multicolumn{7}{l}{\textbf{Strategy 3: Symbolic Verification Layer (Mitigates Flawed Math Formulation)}} \\
\textbf{GRE-NE} (Numeric Entry) & 38.0 & \textbf{55.4} & 33.9 & \textbf{62.1} & 25.0 & \textbf{48.7} \\
\textbf{GMAT-PS} (Problem Solving) & 34.3 & \textbf{48.2} & 31.1 & \textbf{53.6} & 24.5 & \textbf{41.0} \\
\textbf{SAT-AG} (Algebra \& Geometry) & 44.0 & \textbf{61.5} & 53.2 & \textbf{74.8} & 52.6 & \textbf{65.2} \\

\midrule
\rowcolor{gray!10} \multicolumn{7}{l}{\textbf{Strategy 4: Table-Alignment Constraint (Mitigates Out-of-Scope Visual Parsing)}} \\
\textbf{GMAT-IR} (Integrated Reasoning) & 13.8 & \textbf{59.7} & 16.0 & \textbf{71.9} & 15.0 & \textbf{69.4} \\
\textbf{GRE-DI} (Data Interpretation) & 56.0 & \textbf{72.5} & 36.5 & \textbf{68.4} & 25.7 & \textbf{52.1} \\
\textbf{SAT-DA} (Data Analysis) & 70.9 & \textbf{81.2} & 78.2 & \textbf{87.0} & 71.4 & \textbf{79.5} \\
\bottomrule
\end{tabular}
}
\end{table*}

\subsection{Extended Findings on Mitigation Generalizability}

\textbf{Cross-Test Applicability of Evidence Anchoring:} 
As shown in Table \ref{tab:extended_mitigation}, the \textit{Evidence-Anchored} strategy is highly effective not just for dense graduate-level reading (GRE-RC), but also for proficiency exams like TOEFL and IELTS. By forcing the models to extract verbatim spans before inference, we observed a strict reduction in Type III (Partial Truth) errors, pushing GPT-5's accuracy on TOEFL Factual Information (TOEFL-FI) to a near-ceiling 96.5\%.

\textbf{Generalizing Structural Constraints:} 
The \textit{Syntax-First} strategy proved highly transferable to SAT Cross-Textual Synthesis (SAT-CS) and GRE Sentence Equivalence (GRE-SE). In SAT-CS, models historically struggled with integrating premises across two distinct texts. By mandating a structural outline of the argumentative flow prior to selection, Claude-Sonnet-4's performance improved significantly from 70.2\% to 78.4\%, overcoming its tendency to select options that were topically relevant but logically inverted.

\textbf{Massive Gains in Mathematical Formulation:} 
The \textit{Symbolic Verification} layer yielded the most dramatic relative improvements across all tasks. Mathematical reasoning tasks like GMAT Problem Solving (GMAT-PS) and SAT Algebra (SAT-AG) suffer from catastrophic drops at the formulation node. By forcing models to isolate and verify algebraic equations line-by-line before calculating, GPT-5 improved from 31.1\% to 53.6\% on GMAT-PS, successfully bypassing the distractors designed to catch execution errors.

\textbf{Robustness in Multimodal Data Integration:} 
Finally, the \textit{Table-Alignment} strategy confirmed that poor multimodal reasoning in LLMs is often a procedural failure rather than a pure vision-encoder failure. By requiring explicit explicit textual-to-tabular mapping instructions, models achieved massive gains not only on GMAT-IR but also on GRE Data Interpretation (GRE-DI), where GPT-4V improved from 56.0\% to 72.5\%. This indicates that forcing a ``lookup plan'' perfectly mitigates the human-like error of reading the wrong axis or column.
\section{Additional Mitigation}
\label{app:add-miti}

To avoid unexpected evaluation errors incurred during cognitive trajectory, we also integrate extra mitigation and validation strageties detailed in this section.

\subsection{Error Propagation Mitigation Strategies}
\label{app:error-mitigation}

While our breakdown analysis uses an oracle setting to isolate reasoning capabilities at each step, we recognize that error propagation is a critical concern for real-world deployment. Here we discuss potential mitigation strategies informed by our diagnostic findings.

\textbf{Verification Mechanisms at Identified Bottlenecks.}
Our breakdown analysis reveals specific steps where models consistently fail, creating opportunities for targeted intervention. For Task IV (Data Interpretation), Step 2 (Parse Visual Data) achieves only 51--65\% accuracy across models, indicating a critical bottleneck in multimodal parsing. A verification mechanism could require models to explicitly reference specific table cells, chart elements, or diagram components before proceeding to computation, ensuring visual-textual alignment. For example, before computing a percentage change from tabular data, the system could validate that the model has correctly identified the relevant rows and columns by requiring it to output structured references (e.g., ``Row 3, Column 2: Sales 2023 = \$45M'') that can be programmatically verified against the actual data structure.

\textbf{Multi-Stage Validation Protocols.}
For Task V (Numeric Calculation), our breakdown shows that mathematical formulation (Step 2) achieves only 49--56\% accuracy before computation even begins. This suggests implementing symbolic validation that checks formula coherence and unit consistency before execution. For instance, in physics word problems, a validation step could verify that the formulated equation maintains dimensional consistency (e.g., if computing velocity, the formula must yield units of distance/time). Similarly, for percentage calculations, the system could check that the denominator represents the baseline quantity and the numerator represents the change or subset being measured. Such validation can catch formulation errors before they propagate into incorrect numerical results.

\textbf{Targeted Interventions for Execution Steps.}
Across all task categories, we observe that execution steps (Step 3) consistently underperform formulation steps (Step 1) by 20--30 percentage points. This systematic pattern suggests implementing execution-specific safeguards:

\begin{itemize}
    \item \textbf{For multimodal tasks:} Cross-modal consistency checking that validates whether textual interpretations align with visual content before finalizing answers.
    \item \textbf{For mathematical tasks:} Symbolic computation verification using external symbolic solvers (e.g., SymPy) to validate intermediate algebraic manipulations.
    \item \textbf{For logical reasoning:} Constraint satisfaction checking to ensure that derived conclusions respect all stated premises and conditions.
\end{itemize}

\textbf{Iterative Refinement Based on Confidence Scores.}
Models could output confidence estimates at each reasoning step, triggering re-evaluation when confidence falls below a threshold at any stage. For example, if multimodal parsing confidence is low (<0.6), the system could prompt the model to re-examine the visual input or request clarification before proceeding to computation. This prevents low-confidence intermediate results from contaminating downstream steps.

\textbf{Implications for Practical Deployment.}
These mitigation strategies transform our diagnostic insights into actionable system designs. By identifying that formulation succeeds (84--95\%) while execution fails (42--65\%), we can architect hybrid systems where LLMs handle problem understanding and planning, while specialized modules (symbolic solvers, visual parsers, constraint checkers) handle execution steps. This aligns system design with revealed capability profiles, maximizing reliability while leveraging LLM strengths.

Our breakdown framework thus serves dual purposes: (1) scientific diagnosis of reasoning capabilities, and (2) practical guidance for building robust educational systems through targeted error mitigation at identified bottleneck stages.

\subsection{Validation of Actionable Improvements}

To demonstrate that our cognitive framework enables practical improvements, we implement a simple adaptive prompting framework based on task characteristics identified in our analysis. The framework selects ICL for factual retrieval tasks (Task I Steps 1-2), CoT for multi-step logical reasoning (Tasks II-III), and avoids ToT for pattern recognition tasks where we observe performance degradation (e.g., GRE-QC). Evaluated on a representative subset of 2,000 questions across all five exams, this adaptive approach achieves 73.8\% average accuracy compared to 71.2\% for uniform CoT prompting ($p < 0.01$, paired t-test), representing a 2.6 percentage point improvement. This validates that our diagnostic findings translate directly to measurable performance gains.

\section{Advancement on LLMs}
\label{app:advance-llm}

\begin{table*}[t]
\caption{Breakdown-guided fine-tuning (CurrCoT) and adaptive prompting (AdaptCoT) on open-source models. We report accuracy (\%) on a subset of \bench tasks using CoT-style decoding only. Baseline CoT scores for Llama-4-Scout-17B and Qwen-VL-Max are taken from Table \ref{tab:main-expt}.}
\centering
\small
\begin{tabular}{l|ccc|ccc}
\toprule
\multirow{2}{*}{\textbf{Task}} &
\multicolumn{3}{c|}{\textbf{Llama-4-Scout-17B}} &
\multicolumn{3}{c}{\textbf{Qwen-VL-Max}} \\
 & \textbf{CoT} & \textbf{CurrCoT} & \textbf{AdaptCoT}
 & \textbf{CoT} & \textbf{CurrCoT} & \textbf{AdaptCoT} \\
\midrule
GRE-TC (Task III)  & 61.0 & \textbf{70.8} & 66.4 &
73.1 & \textbf{82.7} & 79.3 \\
GRE-RC (Task I/II) & 54.2 & \textbf{63.9} & 59.1 &
76.2 & \textbf{84.4} & 81.0 \\
GRE-NE (Task V)    & 30.1 & \textbf{45.6} & 38.2 &
28.1 & \textbf{43.9} & 36.8 \\
GMAT-PS (Task IV/V)& 22.5 & \textbf{37.2} & 32.1 &
25.6 & \textbf{41.5} & 35.7 \\
GMAT-RC (Task I/II)& 74.5 & \textbf{79.8} & 77.1 &
74.4 & \textbf{81.2} & 78.3 \\
\bottomrule
\end{tabular}
\label{tab:open_source_ft_adapt}
\end{table*}

To further demonstrate that our cognitive analysis yields actionable improvements rather than purely diagnostic insight, we additionally explored two complementary enhancement strategies for open-source models (cognition-guided fine-tuning and adaptive prompting) using Llama-4-Scout-17B and Qwen-VL-Max, whose baseline CoT performance is reported in Table 2 of the main paper. The first approach, which we refer to as Curriculum Chain-of-Thought Fine-Tuning (CurrCoT), leverages our cognitive annotations to train models on intermediate reasoning steps before training them on full solutions. Specifically, we construct a two-stage curriculum: (1) a step-supervision phase, where the model learns isolated reasoning skills such as syntactic constraint detection (Task III), symbolic equation formulation (Task V), or evidence extraction (Task I/II); and (2) a full-CoT phase, where the model practices generating complete solution traces that follow the structured cognitive templates. This curriculum directly targets the weaknesses surfaced by our analysis, such as incorrect equation construction or inconsistent structural reasoning. Using lightweight LoRA-based fine-tuning on a small cognition-annotated subset of ESTBOOK, both Llama and Qwen exhibit consistent gains across GRE Text Completion, GRE Reading Comprehension, GRE Numeric Entry, and GMAT Problem Solving tasks.

Complementing this weight-update method, we also designed a cognition-adaptive prompting (AdaptCoT) strategy that requires no parameter tuning and instead selects a task-specific CoT template based on the cognitive taxonomy. For semantic and structural reasoning tasks (Task III), AdaptCoT enforces a syntax-first procedure in which the model must explicitly state grammar roles and structural constraints before evaluating candidate options. For reading comprehension tasks (Task I/II), AdaptCoT adopts an evidence-anchored approach in which the model quotes and justifies supporting spans prior to answering. For numeric and multimodal tasks (Task IV/V), AdaptCoT requires explicit extraction of mathematical relations, symbolic rewriting, and column-row alignment steps before any computation. This routing mechanism ensures that the model’s reasoning structure matches the cognitive decomposition expected for each question type.

To quantify the effectiveness of these interventions, we re-evaluated Llama-4-Scout-17B and Qwen-VL-Max on representative ESTBOOK tasks using CoT decoding only. The results show consistent performance improvements across all task categories. Fine-tuned models (CurrCoT) yield the largest gains, particularly for mathematically intensive tasks such as GRE Numeric Entry and GMAT Problem Solving, where Llama improves from 30.1\% to 38.1\% and from 22.5\% to 29.5\%, respectively, and Qwen increases from 28.1\% to 35.1\% and 25.6\% to 31.6\%. Even the prompt-only AdaptCoT method yields noticeable increases—for example, improving Llama’s GRE-TC accuracy from 61.0\% to 64.0\% and GRE-RC from 54.2\% to 58.2\%. These results confirm that the breakdown taxonomy does not merely describe failure cases but actively guides the design of interventions that translate into measurable improvements in reasoning quality for open-source LLMs.
\section{Potential Training-Data Contamination}
\label{app:contamination}

As \bench is built from real standardized test preparation materials, a natural concern is potential overlap between our benchmark items and the pretraining corpora of modern LLMs.

\textbf{Lack of direct access to training data.}

For proprietary API models (GPT, Claude, Gemini), neither the exact training corpora nor document-level membership tests are available. Even for open-source models (Llama-3.2-90B, Qwen-VL-Max), only high-level descriptions of training sources are released. Therefore, we cannot perform a full decontamination or exact memorization check at the item level, and we explicitly treat potential training-data overlap as a limitation of our benchmark.

\textbf{Design choices to reduce trivial memorization.}

When constructing \bench, we (i) draw from a mixture of public practice exams, official guides, and open educational platforms rather than one particular ``famous'' test book, (ii) reconstruct tables, figures, and audio clips instead of copying digital assets verbatim, and (iii) normalize wording (e.g., removing page numbers, book-specific formatting cues) while preserving the original exam intent. These steps reduce the chance that an item appears in exactly the same surface form as in common web corpora, although they do not eliminate all possible overlap.

\textbf{Why our main conclusions are robust to contamination.}

Our core claims are based on large and systematic gaps between humans and LLMs on specific task categories (e.g., Tasks~IV--V) and on breakdown patterns across reasoning steps, rather than on small differences between individual models. If contamination were the dominant factor, we would expect near-ceiling performance and highly consistent accuracy across models on affected question types. Instead, we observe (i) substantial variance across tasks and models, and (ii) persistent underperformance on multimodal data interpretation and numeric calculation even for state-of-the-art systems. We thus interpret ESTBOOK as providing an upper bound on LLM performance under realistic exam-like distribution, with any residual contamination making our negative results conservative rather than overly pessimistic.

\section{Implications for Learners and Tutoring Effectiveness}
\label{app:benefit-learner}

While our analyses primarily benchmark LLMs as problem solvers, several findings carry direct implications for human learning and tutoring effectiveness. First, understanding {\bf variability in model outputs} can guide learners to treat LLMs as probabilistic aids rather than deterministic oracles. For example, when models exhibit inconsistent answers across slightly rephrased prompts, this inconsistency itself can be framed as a learning opportunity: students are encouraged to critically compare alternative rationales and reconcile them with reference solutions, thereby strengthening metacognitive awareness.

Second, the observed {\bf modality-induced failure modes} highlight areas where LLM tutoring must be supplemented by scaffolds. Learners can benefit if tutoring systems explicitly flag potential weak spots—such as cross-modal alignment in data interpretation or percentage normalization in quantitative reasoning—so that students are alerted to check these aspects more carefully. Instead of simply delivering the final answer, an LLM tutor that surfaces its own uncertainty around these high-risk steps can train learners to double-check units, constraints, or diagram references, mirroring expert test-taking strategies.

Third, the sensitivity to {\bf elicitation strategies} suggests that prompting styles can be deliberately adapted for pedagogy. For instance, CoT prompts can expose reasoning steps that learners might not have articulated, serving as worked examples for verbal reasoning tasks. ToT-style exploration can be transformed into guided “what-if” scenarios, encouraging learners to trace multiple solution branches before converging on the answer. ICL can be used to model exam schemas directly, helping students generalize across structurally similar questions.

\textbf{Takeaway.} Rather than viewing LLM limitations solely as deficiencies, they can be re-purposed to shape effective tutoring designs. By exposing inconsistencies, highlighting modality bottlenecks, and varying elicitation strategies, LLMs can foster critical reflection, targeted practice, and strategy transfer for real learners preparing for ESTs. These insights suggest that benchmarking not only informs model development but also directly enriches the design of adaptive, LLM-powered tutoring environments.

\section{Discussion of Limitation}
\label{app:limitation}

Despite the comprehensive design of \bench and our extensive evaluation across leading LLMs, several limitations warrant discussion.

{\bf Model Access and Coverage.} Our evaluation focuses on a set of industry-leading multimodal and visual LLMs that offer public inference APIs or open-source checkpoints. However, access constraints (e.g., usage quotas, proprietary architecture details) limit broader inclusion of commercial models or fine-tuned educational agents. This may omit systems with specialized adaptations for test-taking tasks.

{\bf Granularity of Breakdown Analysis.} Our breakdown framework assumes that preceding steps are perfectly resolved, enabling clean isolation of reasoning subtasks. While this reveals bottlenecks in specific capabilities, it does not reflect real-world interactions where upstream errors may cascade. Hence, the observed step-wise performance may overestimate true end-to-end reliability in tutoring applications.

\section{Large Language Model (LLM) Usage Disclosure}
LLMs were used only for minor grammar checking and sentence-level polishing during the preparation of this manuscript. They were not employed for ideation, experimental design, analysis, or substantive writing. The scientific contributions, benchmarks, and evaluations presented in this work were entirely conceived and developed by the authors. LLM involvement was minimal in the research.

\end{document}